\documentclass[10pt,twocolumn,letterpaper]{article}

\usepackage{mpi}
\usepackage{times}
\usepackage{epsfig}
\usepackage{graphicx}
\usepackage{amsmath}
\usepackage{amssymb}
\usepackage{multirow}
\usepackage{booktabs}
\usepackage{mathtools}

\usepackage{caption}
\usepackage{subcaption}
\usepackage{algpseudocode,algorithm,algorithmicx}
\usepackage{xr}
\usepackage{bm}
\usepackage{color}
\usepackage{makecell} 

\def\bfR{{\bf R}}
\def\bfS{{\bf S}}
\def\bfW{{\bf W}}

\def\bfA{{\bf A}}

\def\bfI{{\bf I}}

\def\bfD{{\bf D}}
\def\bfE{{\bf E}}

\def\so3{{ \text{SO(3)}}}

\def\bfq{{\bf q}}

\newcommand{\norm}[1]{\left\lVert#1\right\rVert}

\usepackage[font=footnotesize, labelfont=footnotesize, labelfont=bf]{caption} 
\usepackage{cuted}
\usepackage{capt-of}
\usepackage[pagebackref=true,breaklinks=true,letterpaper=true,colorlinks,bookmarks=false]{hyperref}

\mpifinalcopy

\begin{document}

\title{Intrinsic Dynamic Shape Prior for Fast, Sequential and Dense Non-Rigid Structure from Motion with Detection of Temporally-Disjoint Rigidity\thanks{$\,$supported by the ERC Consolidator Grant 4DReply (770784) and the BMBF projects DYNAMICS (01IW15003) and VIDETE (01IW18002).}\vspace{-5pt}} %

\author{Vladislav Golyanik$^1$\hspace{2.3em}
Andr\'{e} Jonas$^2$\hspace{2.35em}
Didier Stricker$^{2,3}$\hspace{2.3em}
Christian Theobalt$^1$\vspace{0.3em}\\
\hspace{-55pt}
$^{1}$MPI for Informatics\hspace{2.4em}
$^{2}$University of Kaiserslautern\hspace{2.59em}
$^{3}$DFKI
}

\maketitle

\begin{abstract} 
While dense non-rigid structure from motion (NRSfM) has been extensively studied from the perspective of the reconstructability problem over the recent years, almost no attempts have been undertaken to bring it into the practical realm. 
The reasons for the slow dissemination are the severe ill-posedness, high sensitivity to motion and deformation cues and the difficulty to obtain reliable point tracks in the vast majority of practical scenarios. 
To fill this gap, we propose a hybrid approach that extracts prior shape knowledge from an input sequence with NRSfM and uses it as a dynamic shape prior for sequential surface recovery in scenarios with recurrence. %
Our \textit{Dynamic Shape Prior Reconstruction} (DSPR) method can be combined with existing dense NRSfM techniques while 
its energy functional is optimised with multi-start gradient descent at real-time rates for new incoming point tracks. 

The proposed versatile framework with a new core NRSfM approach outperforms several other methods in the ability to handle inaccurate and noisy point tracks, provided we have access to a representative (in terms of the deformation variety) image sequence. 
Comprehensive experiments highlight convergence properties and the accuracy of DSPR under different disturbing effects. 
We also perform a joint study of tracking and reconstruction and show applications to shape compression and heart reconstruction under occlusions. 
We achieve state-of-the-art metrics (accuracy and compression ratios) in different scenarios. 
\end{abstract} 

\vspace{-7pt} 

\section{Introduction}\label{sec:intro_and_contributions}

Dynamic non-rigid 3D reconstruction from monocular image sequences relying exclusively on motion and deformation cues and weak prior assumptions is known as \textit{non-rigid structure from motion} (NRSfM) \cite{Bregler2000, Brand2005, Torresani2008}. Despite advances over recent years in the reconstruction accuracy and variety of scenarios which can be handled by NRSfM \cite{Rabaud2008, Gotardo2011, Garg2013, Golyanik2017_SPVA, Kumar2017}, there is a gap between results achieved in a controlled environment and real scenarios. 
Often, it is difficult to obtain reliable dense correspondences across input views. 
Due to the high ill-posedness of NRSfM, there is no universal set of prior constraints that works equally well across different scenarios. 

The \textbf{main contribution} of this paper is a new fast and sequential technique for dense monocular non-rigid reconstruction with a dynamic shape prior (DSP), \textit{i.e.,} a sequence-specific set of ordered and gradually changing 3D states obtained on a representative image sequence (Sec.~\ref{sec:approach}). %
In the vast majority of real-world cases, not deformations but rather different angles of view (camera poses) cause different 2D measurements. 
It is assumed that the representative sequence provides a sufficient variety of deformations as they are likely to occur in a given scene, whereas there are no strong requirements for poses except that those must be nondegenerate. 
While the DSP generation is offline, the reconstruction of new frames with DSP is light-weight and well parallelisable. 
It implicitly assumes temporally-disjoint rigidity, \textit{i.e.,} the situation when a newly observed 3D state is reoccurring with respect to the DSP. 

For every new incoming measurement, the proposed shape-from-DSP or \textit{Dynamic Shape Prior Reconstruction} (DSPR) approach finds a globally optimal 3D state corresponding to the 2D measurements and rigidly transforms it to the pose as observed in the measurements by alternating between multi-start gradient descent (MSGD) and camera pose estimation. Note that the pose in the incoming frames can be arbitrary and differ significantly from poses observed during the DSP generation in the generative sequence, due to the decoupling property of shapes and poses in NRSfM. 
Thus, our framework can be considered as a variant of incremental NRSfM, since we decouple the basis estimation from the weights and camera poses. See Fig.~\ref{fig:HEART_SEQUENCE} for an example of monocular non-rigid reconstruction with DSPR. 
As a \textbf{further contribution}, we propose a new light-weight dense per-point extension of \cite{Golyanik_2019} which we call \textit{Dense Consolidating Monocular Dynamic Reconstruction} (D-CMDR) approach for the DSP recovery from a representative sequence, even though any accurate existing dense NRSfM method can be employed for this task (Sec.~\ref{sec:obtaining_shape_prior}). %
Thus, \textit{the focus of this paper is towards making NRSfM applicable in real-world scenarios}, and not improving the accuracy of NRSfM from the perspective of the reconstructability problem \textit{per se}. 
Apart from real-time monocular reconstruction from noisy data, our main idea 
can also be applied to several related problems. 
Since DSP represents a compact footprint of the geometry carrying a learned sequence-specific deformation model, it suggests the suitability of DSPR for \textbf{geometry compression} (Sec.~\ref{ssec:experiments_with_real_data}). 
We thoroughly evaluate our DSPR framework and the D-CMDR approach for the DSP generation (Sec.~\ref{sec:experiments}). 
Apart from the standard NRSfM datasets and the NRSfM challenge \cite{Jensen2018} covering more than fifteen methods (Sec.~\ref{ssec:CMDR_disjointly}), we synthesise a new \textit{actor mocap} dataset for joint evaluation of dense point tracking and reconstruction (Sec.~\ref{ssec:joint_evaluation}). 
Moreover, compared to the prevalent evaluation policy of dense NRSfM in the literature, we evaluate our framework with perturbed point tracks and missing data (Sec.~\ref{subs:main_experiment}). 

\section{Related Work}\label{sec:related_work}

  Some recent works on NRSfM focus on dense \cite{Garg2013} and scalable methods \cite{Ansari2017, Kumar2018, Kumar2019} as well as approaches for complex non-linear deformations \cite{Zhu2014, Kong_2016}. 
  A distinct tendency is investigating new, often simple and, at the same time, overlooked ideas \cite{Dai2017, Li2018, Kovalenko2019} and models for NRSfM \cite{Agudo_Moreno_Noguer_2015, Golyanik2017}. 
  More attention is paid to hybrid methods which make stronger assumptions than classic NRSfM but fewer assumptions than template-based counterparts or domain-specific approaches which expect a known object class \cite{Blanz1999, Tewari2017}. 
  One example of hybrid techniques is an approach for handling occlusions with a static shape prior obtained on several unoccluded frames of a sequence \cite{Golyanik2017_SPVA}. 
  Some methods with a trained deformation model rely on a representative dataset for training \cite{Pumarola2018, Shimada_2019}. 
  Our algorithm has thrived on the ideas proposed in the works mentioned above. 
  The most closely related methods to DSPR are~\cite{Li2018} and \cite{Golyanik2017_SPVA}. 
  \vspace{2 pt} \\ 
  \textbf{The method of Li \textit{et al.}~\cite{Li2018}.} 
  Li \textit{et al.}~\cite{Li2018} propose to exploit state recurrency in sparse NRSfM. 
  While a local rigidity method rapidly reaches its lower bound on the number of views necessary for the rigid reconstruction to produce meaningful results \cite{Rehan2014}, the method of Li \textit{et al.}~\cite{Li2018} does not rely on connected temporal windows and is agnostic to the deformation intensity over a short period. 
  The number of rigid clusters has to be set in \cite{Li2018} in advance. 
  Besides, if some states are unique or degenerate (are not observed in other poses), they are assigned to some non-empty clusters and treated as noise. 
  Thus, non-reoccurring states are reconstructed less accurately. 
  Moreover, the method of Li \textit{et al.}~\cite{Li2018} requires computationally costly graph clustering and works for a few sparse points. 
  In contrast, DSPR fits an instance from DSP which is related to a given dense 2D measurement by rigidity. 
  We do not explicitly cluster dense point tracks into bins relating the underlying 3D states by rigidity. 
  Instead, we find a subsequence providing deformations as diverse as possible in as few views as possible. 
  \vspace{2 pt} \\ 
  \textbf{Shape Priors and Degenerate Data Handling.} Del Bue~\cite{DelBue2008} proposed an NRSfM factorisation with a supportive pre-computed shape basis. 
  The method was shown to handle degeneracies in the sparse point tracks robustly. 
  Golyanik \textit{et al.}~\cite{Golyanik2017_SPVA} included a static shape prior into dense variational NRSfM. %
  Compared to them, we extract multiple states from a representative sequence reflecting the entire deformation model. 
  While the aim of~\cite{Golyanik2017_SPVA} is the stability under large occlusions, their method also tends to overconstrain the reconstructions. 
  Our primary goal is a light-weight sequential scheme with recurrent state identification, and still, it is remarkably robust under occlusions. 

  Several methods for sparse NRSfM address missing data \cite{OlsenBartoli2008, GotardoMartinez2011, Gotardo2011, Agudo_Moreno_Noguer_2015, Lee2017}. 
  Gotardo and Martinez~\cite{GotardoMartinez2011, Gotardo2011} rely on a pre-defined trajectory basis and the smooth deformations constraint while recovering a low-rank approximation of the measurement matrix with estimated missing entries. 
  The approach of Lee \textit{et al.}~\cite{Lee2017} is robust to moderate portions of missing data as the shape likelihoods are influenced only by available entries in their method. 
  Our DSPR approach is robust to moderate portions of missing entries. Note that we treat those as erroneous measurements, which is a more realistic assumption in the dense setting. 
  \vspace{2 pt} \\ 
  \textbf{Sequential NRSfM}. The majority of NRSfM methods operate globally on frame batches \cite{Paladini2012, Garg2013, Zhu2014, Agudo_Moreno_Noguer_2015, Kong_2016, Ansari2017, Kumar2018}. 
  Paladini~\textit{et al.}~\cite{Paladini2010} proposed the seminal sequential method which incrementally updates deformation modes upon the data availability. 
  Agudo and coworkers~\cite{Agudo2014} introduced a probabilistic model with physics-based constraints for dense sequential NRSfM. 
  Once DSP is obtained, our DSPR switches to the sequential reconstruction and requires only a single measurement and the latest regressed 
  surface as an input. 
  In contrast to \cite{Paladini2010, Agudo2014}, it is explicitly designed with the handling of inaccurate correspondences in mind. 
  Moreover, our optimisation is very fast and highly parallelisable. 
  It is possibly faster than most of the NRSfM algorithms in the literature so far, considering the methods \cite{Agudo2014, AMP2017, Agudo2018}. \\ 
  \textbf{Recovery of the Dynamic Shape Prior.} In the proposed D-CMDR for the reconstruction of a representative sequence, up to several millions of parameters are optimised with non-linear least squares (NLLS). 
  D-CMDR is tailored for the dense per-point case and is a variant of the segmentwise CMDR \cite{Golyanik_2019}. 
  The most closely related approach to D-CMDR is the template-based method of Yu \textit{et al.}~\cite{Yu_2015}, with several differences: 
  1) instead of using a multiview reconstruction to obtain a template, we initialise shapes and camera poses with the rigid factorisation \cite{TomasiKanade1992}; 
  2) we use trajectory regularisation instead of as-rigid-as-possible regulariser \cite{SorkineAlexa2007}, and 3) the fitting term operates on point tracks and not directly on images. 
  An NRSfM technique with simultaneous constraints in metric and trajectory spaces is Column Space Fitting \cite{Gotardo2011}. 
  Our trajectory smoothness term was rarely used in energy-based NRSfM so far. 
  It allows integration of subspace constraints on point trajectories and originates from \cite{Akhter2011}. 
  We demand smoothness of neighbouring trajectories by optimising the total variation of trajectory 
  coefficients. 
  A similar regulariser was previously applied in multi-frame optical flow (MFOF) \cite{Garg2013flag, Taetz2016}. 
  Olsen and Bartoli \cite{OlsenBartoli2008} proposed one of the first spatial regularisers with a related principle, \textit{i.e.,} a surface continuity prior term imposing similarity constraint on neighbouring point trajectories for the enhanced robustness against missing data. 

\section{The Proposed DSPR Approach} 
\label{sec:approach} 

Our objective is the 3D reconstruction of a current 3D state $\bfS_f \in \mathbb{R}^{3 \times N}$ given incoming measurements $\bfW_f \in \mathbb{R}^{2 \times N}$, $f \in \{1, \hdots, F\}$ and a DSP $\bfD = \{ \bfD_i \}$, $i \in \{1, \hdots, Q\}$ with $Q$ temporal rigidity bases. 
$F$ is the total number of frames and $N$ is the number of points per frame. 
We formulate dense sequential NRSfM as a per-frame energy minimisation problem of finding $\bfD_i$ related to $\bfS_f$ by a rigid transformation and camera pose $\bfR_f \in \mathbb{R}^{3 \times 3}$ ($\bfR_f^\mathsf{T} = \bfR_f^{-1}$, $\operatorname{det}(\bfR_f) = 1$) so that the product $\bfR_f \bfD_i$ explains the current observation $\bfW_f$: 
\begin{equation}\label{eq:DSPR_SEARCH} 
\begin{aligned} %
  & \bfE(\bfS_f = \bfD_i, \bfR_f) =  \alpha \norm{\bfW_f - \bfI_{2  \times 3} \bfR_f \, \bfD_{i: \lambda_i = 1}}_\mathcal{F} +  
  \\ & \;\;\;\;\;\;\;\;\; \beta \norm{\bfD_{i: \lambda_i = 1} - \bfS_{f-1}}_\mathcal{F}  
   + \gamma \, ( \norm{\boldsymbol{\lambda}}_{0} - 1)^2, 
\end{aligned} 
\end{equation} 
where $\norm{\cdot}_0$ and $\norm{\cdot}_\mathcal{F}$ denote a zero-norm of a vector and Frobenius norm, respectively, $\bfI_{2  \times 3}$ models orthographic projection and $\boldsymbol{\lambda} = [\lambda_i]$ is the indicator function for DSP. 
The energy functional~\eqref{eq:DSPR_SEARCH} contains a data term weighted by $\alpha$, temporal smoothness term weighted by $\beta$ and a DSP regularisation term weighted by $\gamma$. %
The data term ensures that the factorisation $\bfR_f \bfS_f$ is accurately projected to $\bfW_f$. 
The smoothness term expresses the assumption of the gradual character of changes in the states as well as helps to converge faster. 
The regulariser ensures that a single $\bfD_i$ is required to explain observations upon our model. 
This practice contrasts to some other methods, where every shape is encoded as a linear combination of basis shapes (recovered during the reconstruction or known in advance) \cite{Paladini2012}. 
In our model, DSP is assumed to provide a sufficient variety to cover the entire space of reoccurring deformations, and we use the decoupling property of the shape and pose. 

The energy functional~\eqref{eq:DSPR_SEARCH} is minimised iteratively, by alternatingly fixing $\bfR_f$ and releasing $\bfD_{i: \lambda_i = 1} = \bfS_f$, and vice versa, in every iteration. 
When $\bfS_f$ is fixed, the only term dependent on $\bfR_f$ is the data term. $\bfR_f$ can be updated in the closed-form by projecting its affine update to the $SO(3)$ group or by linear least squares with quaternion parametrisation. 
When $\bfR_f$ is fixed, an optimal $\bfD_i$ is found by taking the partial derivative of the energy subspace with the fixed $\bfR_f$, denoted by $\bfE_{\bfR_f}$, 
\hbox{w.r.t.}~$\boldsymbol{\lambda}$ and equating it to zero: 
\begin{align}\label{eq:optimality_criterion} 
  \frac{\partial \bfE_{\bfR_f}(\bfS)}{\partial \bfS} \frac{ \partial  \bfS }{\partial \boldsymbol{\lambda} } = 0. 
\end{align} 
The optimality criterion in Eq.~\eqref{eq:optimality_criterion} defines a state when a small change in the shape caused by a small 
change in the prior state does not change the energy. 
We minimise the energy functional~\eqref{eq:DSPR_SEARCH} --- when $\bfR_f$ is fixed --- by the \textit{multi-start gradient descent} (MSGD) method. 
Starting from multiple regularly sampled values of $\boldsymbol{\lambda}$, we compute differences in $\bfE_{\bfR_f}$ and update $\boldsymbol{\lambda}$ %
in the direction of the energy decrease. Multiple starting points are required to obtain a globally optimal solution since $\bfE$ is non-convex. 
The global minimum is obtained by comparing locally minimal energy values. 
MSGD is well parallelisable as every thread can converge or finish upon a boundary condition (\textit{e.g.,} when leaving the assigned range of values) independently from other threads. 
Thanks to MSGD, DSPR executes with three-five frames per second on our hardware without parallelisation (see Sec.~\ref{sec:experiments}). 

\subsection{Obtaining Dynamic Shape Prior (DSP)}\label{sec:obtaining_shape_prior}

  DSP generation includes an accurate 3D reconstruction of a representative image sequence with a general-purpose NRSfM method. 
  In principle, we are free to choose any dense scalable NRSfM technique for the initial reconstruction. In the quantitative experiments, %
  we use two accurate existing methods, \textit{i.e.,} Garg \textit{et al.} \cite{Garg2013} and Ansari \textit{et al.} \cite{Ansari2017}. 
  Additionally, we propose a new energy-based NRSfM method which outperforms the approaches mentioned above in a subset of evaluation scenarios. %

  \subsubsection{Our Core NRSfM Approach for DSP Acquisition}\label{ssec:CMDR}

  For notational consistency, we denote the measurements of the representative sequence and the corresponding 3D shapes in this section by $\bfW_{2F \times N} = [\bfW_f]$ and $\bfS_{3F \times N} = [\bfS_f]$ respectively, with $N$ denoting the number of points in every frame. 
  The new method minimises the following energy functional with the Gauss-Newton algorithm: 
  \begin{equation}\label{eq:CMDR} 
  \begin{aligned} 
    & \bfE_{\text{D-CMDR}}(\bfR, \bfS, \bfA) = \alpha \, \bfE_{\text{fit}}(\bfR, \bfS) + \beta \, \bfE_{\text{temp}}(\bfS) + \\
    & \;\;\;\;\;\;\;\;\;\; + \lambda \, \bfE_{\text{linking}}(\bfS, \bfA) + \rho \, \bfE_{\text{reg}}(\bfA), 
  \end{aligned} 
  \end{equation} 
  where $\bfA$ is a matrix with trajectory coefficients 
  explained below. 
  The data term constrains projections of the recovered shapes to agree with the 2D measurements: 
  \begin{align}
    \bfE_{\text{fit}}(\bfR, \bfS) = \sum_f \norm{\bfW_f -  \bfI_{2 \times 3} \bfR_f \, \bfS_f }_{\epsilon}^{2}, 
  \end{align} 
  where $\norm{\cdot}_{\epsilon}$ is Huber norm ($\epsilon = 0.1$). %
  The temporal smoothness term imposes similarity on adjacent reconstructions: %
  \begin{equation}
    \bfE_{\text{temp}}(\bfS) = \sum_{f = 2}^{F} \norm{ \bfS_{f} - \bfS_{f - 1}  }_{\epsilon}^{2}. 
  \end{equation}
  The linking term expresses our assumptions about the complexity of deformations (deformation model). 
  Here, we rely on $K$ known basis trajectories $\Theta$ sampled from discrete cosine transform (DCT) at regular intervals: 
  \begin{equation}\label{eq:linking} 
    \bfE_{\text{linking}}(\bfS, \bfA) = \norm{ \bfS
                                        - 
                                        (\Theta \otimes \bfI_{3})_{3F \times 3K} \, \bfA_{3K \times N}
                                        }_{\epsilon}^{2}, \;\text{where} 
  \end{equation}
  \begin{equation}
  \begin{aligned}
    & \Theta  =
	\begin{pmatrix} 
		[\theta_{11} & \hdots & \theta_{1K}] & \hdots & [\theta_{F1} & \hdots & \theta_{FK}]
	\end{pmatrix}^{\mathsf{T}}, \\ %
	& \;\;\;\;\;\;\;\;\; \theta_{tk}  = \frac{\sigma_{k}}{\sqrt{2}} \cos \left(\frac{\pi}{2F} (2t - 1)(k - 1) \right) \text{and} \\
	& \;\;\;\;\;\;\;\;\; \;\;\;\;\;\;\;\;\; \sigma_{k}  = 
	\begin{cases}
		1 & \text{if } k = 1 \text{,}\\
		\sqrt{2} & \text{otherwise.} 
	\end{cases}
  \end{aligned}
  \end{equation}
  In Eq.~\eqref{eq:linking}, $\bfA$ holds coefficients of linear combinations which approximate reconstructed 3D trajectories. %
  $\bfE_{\text{linking}}$
  connects or \textit{links} these trajectories to unknown though valid combinations of basis trajectories. %
  Depending on the linking strength, the calculated trajectories will more or less accurately resemble valid combinations of basis trajectories. 
  Finally, the regularisation term imposes a temporal coherence constraint on 3D trajectories of adjacent points. %
  Since the recovered 3D trajectories are parameterised by $\bfA_k$, the regularisation term can be expressed as 
  \vspace{-3pt}
  \begin{equation}\label{eq:regularisation} 
  \vspace{-1pt} 
    \bfE_{\text{reg}}(\bfA) = \sum_{n = 1}^{N} \sum_{k = 1}^{K}  \norm{ \nabla \bfA_{k, n} }_\epsilon^{2}. 
  \end{equation} 
  To calculate gradients of trajectory coefficients, Eq.~\eqref{eq:regularisation} requires a point adjacency lookup table which is derived from the spatial arrangement of the points in the reference frame. % 
  Our core NRSfM approach is called \textit{Dense Consolidating Monocular Dynamic Reconstruction} (D-CMDR), as it unifies constraints in the metric 
  and trajectory spaces into a single energy functional. %
  In the beginning, $\bfS$ and $\bfR$ are initialised under rigidity assumption with \cite{TomasiKanade1992} on the unaltered point tracks $\bfW$. 
  $\alpha$, $\lambda$ and $\rho$ are usually equivalued, while $\beta$ is set an order of magnitude lower. 

  \subsubsection{Postprocessing of DSP} 
  
  After the reconstruction of the representative sequence, 
  we obtain $L$ shapes $\bfS^\sharp_l$, $l \in \{1, \hdots, L \}$. 
  The recovered poses are not applied to $\bfS^\sharp_l$ and discarded but a single global arbitrary pose for all $\bfS^\sharp_l$ can be chosen. 
  Next, we build a map of pairs $\chi = (|| \bfS^\sharp_l ||_\mathcal{F}, \bfS^\sharp_l)$ where the shapes are arranged %
  in the increasing order of 
  $|| \bfS^\sharp_l ||_\mathcal{F}$. Starting from $\bfS^\sharp_1$, we iteratively include $\bfS^\sharp_l$ into DSP 
  if the norm difference between the current $\bfS^\sharp_l$ and the latest included $\bfD_i$ exceeds some $\mu$. 
  By varying $\mu$, we can control $Q$, \textit{i.e.,} the cardinality of $\bfD$. Experimentally, we observe %
  a strong correlation between $|| \bfS^\sharp_l ||_\mathcal{F}$ values and the corresponding shapes, \textit{i.e.,} if Frobenius norms %
  are similar, the shapes are close likewise\footnote{if shapes are related by reflection around the $yz$-plane (which is rarely the case in practice), they will have the same Frobenius norm}. 

\section{Experimental Evaluation} 
\label{sec:experiments} 

This section outlines the evaluation methodology and summarises the results. 
We implement DSPR in C++ for a single thread. 
All values are reported for a system with 32 Gb RAM and Intel Core i7-6700K processor with cores running at 4.00GHz under Ubuntu 16.04.3.

\subsection{Evaluation Methodology}

We develop several tests with synthetic and real data for the evaluation of the convergence, accuracy and runtime aspects of DSPR. 
For the DSP reconstruction, we use several NRSfM methods based on different principles , \textit{i.e.,} Variational Approach (VA) \cite{Garg2013}, Scalable Monocular Surface Reconstruction (SMSR) \cite{Ansari2017} and the proposed D-CMDR. 
Depending on the evaluation scenario, we report different metrics characterising the accuracy of geometry and camera pose estimation. 
Let $\bfS'_f$ and $\bfR'_f$, $f \in \{1, \hdots, F \}$, be the ground truth geometries and camera poses respectively. 
As a shape fidelity metric, we report a \textit{mean root-mean-square error} (RMSE) for a set of views 
defined as $e_{3D} = \frac{1}{F} \sum_{f} \frac{||\bfS'_f - \bfS_f||_{\mathcal{F}}}{||\bfS'_f||_{\mathcal{F}}}$, 
where $\norm{\cdot}_{\mathcal{F}}$ is Frobenius norm. 
Since the camera poses are recovered up to an arbitrary rotation, we find a single optimal corrective rotation $\bfR^{\sharp}$ aligning the recovered poses and the ground truth camera poses. 
Thus, for the evaluation purposes we solve the energy minimisation problem  $\min_{\bfR^{\sharp}} \, \sum_{f} \norm{\bfR'_{f} - \bfR^{\sharp} \, \bfR_{f} }_{\epsilon}$, 
with $\norm{\cdot}_{\epsilon}$ denoting Huber norm with the threshold value $\epsilon = 1.0$. 
After applying $\bfR^{\sharp}$ to all $\bfR_{f}$, we compute a \textit{mean quaternionic error} (QE) defined as $e_{q} = \frac{1}{F} \sum_{f} |\bfq'_f - \bfq_f |$, 
with $|\cdot|$ standing for the quaternion norm. $\bfq'_f$ and $\bfq_f$ are the quaternions\footnote{here, the quaternions are guaranteed to have a positive sign} corresponding to $\bfR'_{f}$ and $\bfR^{\sharp} \bfR_{f}$ respectively. 

\begin{table}[!t] 
  \scriptsize 
  \begin{center}
    \begin{tabular}{|p{14pt}|p{14pt}|p{14pt}|p{14pt}|p{14pt}|p{14pt}|p{14pt}|p{14pt}|p{17pt}|} \hline %
	 \textbf{TB} \cite{Akhter2011} & \textbf{MP} \cite{Paladini2012} & \textbf{VA} \cite{Garg2013} & \textbf{DSTA} \cite{Dai2017} & \textbf{CDF} \cite{Golyanik2017} & \textbf{SMSR} \cite{Ansari2017} & \textbf{GM} \cite{Kumar2018} & \textbf{JM} \cite{Kumar2019} & \textbf{CMDR (ours)}  \\ \hline\hline %
	0.1252		&	0.0611		&	0.0346	     &		0.0374	   & 	0.0886	& 	0.0304		  &  0.0294 & 0.280 & 0.0324   	\\ \hline	
	0.1348		&	0.0762 		& 	0.0379	     & 		0.0428	   & 	0.0905	& 	0.0319		  &  0.0309 & 0.327 & 0.0369	\\ \hline
  \end{tabular}
  \end{center}
  \vspace{-10pt}
  \caption{Mean RMSE on \textit{seq.~A} (the first row) and \textit{seq.~B} (the second row).} %
  \label{tab:comparison_synth_faces} 
\end{table}

\begin{figure*}[t!] 
\centering 
  \includegraphics[width=1.0\linewidth]{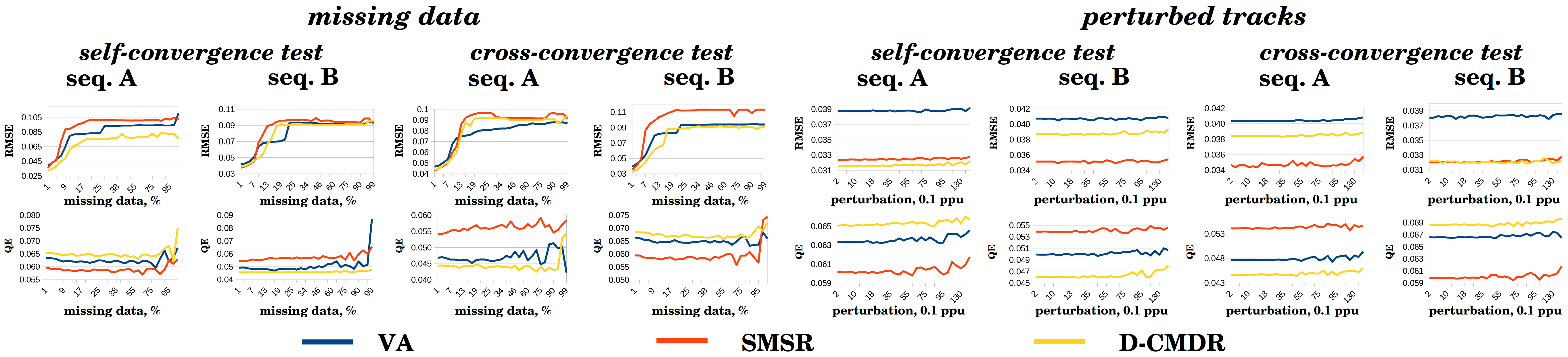} %
  \caption{ Experimental results of the \textit{self-} and \textit{cross-convergence} tests with missing and perturbed data. 
  Three core approaches are tested, \textit{i.e.,} VA \cite{Garg2013}, SMSR \cite{Ansari2017} and D-CMDR (ours). In all experiments, we report mean RMSE
  and QE as the functions of missing data ratio (in $\%$) and perturbation (measured in $0.1$ pixels per unit or \textit{ppu}). 
  Missing data is varied in the range $[0; 99]\%$, and the perturbation is varied in the range $[0; 15]$ pixels. 
  } 
  \label{fig:STATISTICS_2} 
\end{figure*} 

\begin{table*}[!htb]
  \scriptsize
  \begin{center}
   \begin{tabular}{|c|c||c|c|c|c|c|c|c||c|c|c|c|c|} \hline %
   \multicolumn{2}{|c||}{\multirow{2}{*}{\textbf{METHOD}}} &\multicolumn{7}{c||}{\textbf{PERTURBED DATA}} & \multicolumn{5}{c|}{\textbf{MISSING DATA}} \\\cline{3-14}
   \multicolumn{2}{|c||}{}			  & $\bm{0.4}$ \textbf{px}   & $\bm{1.2}$ \textbf{px}  	& $\bm{1.6}$ \textbf{px} 	& $\bm{2.0}$ \textbf{px} 	& $\bm{3.0}$ \textbf{px} & $\bm{4.0}$ \textbf{px} & $\bm{5.0}$ \textbf{px} 	 & $\bm{1\%}$   	& $\bm{3\%}$         	         	& $\bm{11\%}$ 	  	& $\bm{17\%}$       	& $\bm{23\%}$    \\\hline 		%
   \multirow{2}{*}{SMSR \cite{Ansari2017}}  & RMSE	  &    0.0455    	     & 0.0962 			& 0.1243  			& 0.1536 			&  0.2232 		 & 0.2956 		  & 0.3885 	 		 & 0.1001 		& 0.1778      		     		& 0.3365  		& 0.4143     		& 0.4849   	\\\cline{2-14} 		%
				  & QE	  &    0.2434    	     & 0.2999   		& 0.3287 			& 0.2450 			&  0.3068 		 & 0.2280 		  & 0.3510 			 & 0.2972 		& 0.2973     		    		& 0.2968  		& 0.2968     		& 0.2975  	\\\hline 		%
    \multirow{2}{*}{D-CMDR (ours)} & RMSE	  &    0.0646    	     & 0.1918 			& 0.2541  			& 0.2867			&  0.3571		 & 0.4056 		  & 0.4522 	 		 & 0.1001 		& 0.1777      		      		& 0.3365 		& 0.4143		& 0.4849	\\\cline{2-14} 		%
				  & QE	  &    0.0689    	     & 0.1077   		& 0.1514			& 0.1711			&  0.4617 		 & 0.4578 		  & 0.4506			 & 0.0663 		& 0.0663    		   		& 0.0662		& 0.0663		& 0.0663	\\\hline 	%
   \multirow{2}{*}{DSPR (ours)}   & RMSE	  &    \textbf{0.0324}    	     & \textbf{0.0324} 			& \textbf{0.0324}  			& \textbf{0.0324} 			&  \textbf{0.0324} 		 & \textbf{0.0324} 		  & \textbf{0.0325} 	 		 & \textbf{0.0327} 		& \textbf{0.03578}     & \textbf{0.0754}  		& \textbf{0.0962}    		& \textbf{0.0994}   	\\\cline{2-14} 		%
				  & QE	  &    \textbf{0.0602} & \textbf{0.0601}   & \textbf{0.0600} 	& \textbf{0.0601} &  \textbf{0.0603} 		 & \textbf{0.0600}   & \textbf{0.0603} & \textbf{0.0602} 	& \textbf{0.05984}     & \textbf{0.0584}  & \textbf{0.0585}     		& \textbf{0.0581}  	\\\hline 		%
   \end{tabular} 
\end{center}
\vspace{-12pt} 
\caption{Mean RMSE and mean QE for SMSR \cite{Ansari2017} and D-CMDR (our method for DSP reconstruction) on perturbed tracks and tracks with missing entries. } 
\label{table:comparisons_other_methods} 
\end{table*}

Next, we evaluate the core D-CMDR approach individually (Sec.~\ref{ssec:CMDR_disjointly}) and jointly with DSPR (Secs.~\ref{subs:main_experiment}-\ref{ssec:SGD_parameters}). 
We perform self- and cross-convergence tests of DSPR with perturbed and missing data (Sec.~\ref{subs:main_experiment}), MSGD convergence tests (Sec.~\ref{ssec:SGD_parameters}) and joint evaluation of flow and DSRP (Sec.~\ref{ssec:joint_evaluation}). In the \textit{self-convergence test,} DSP is reconstructed on ground truth point tracks, and the same tracks are used for the evaluation, whereas in the \textit{cross-convergence test}, the point tracks for the reconstruction of the shape prior and the DSP are different. 
\begin{samepage}
In Secs.~\ref{ssec:CMDR_disjointly}--\ref{ssec:SGD_parameters}, we use two $99$ frames long synthetic face sequences with known geometry and dense point tracks from \cite{Garg2013}. Both sequences $\;A\;\,$and$\;\,B\;$originate$\;\,$from$\;\,$the$\;\,$same$\;\,$set$\;\,$of$\;\,$facial 
\end{samepage}

\hspace{-16pt} expressions. The difference lies in the series of camera poses applied to the interpolated expressions. 
Due to the different camera pose patterns, the sequences are of varying difficulty for NRSfM and reconstructed differently (in many cases close to each other but not exactly in the same way). 
Thus, they offer an optimal testbed for the cross-convergence test. 
Finally, we show applications of DSPR on real data and report shape compression ratios (Sec.~\ref{ssec:experiments_with_real_data}). 

\subsection{Evaluation of D-CMDR Separately from DSPR}\label{ssec:CMDR_disjointly} 

Although D-CMDR is evaluated jointly with DSPR in the following, we report mean RMSE for it on synthetic faces, see Table~\ref{tab:comparison_synth_faces}. 
The errors for Trajectory Basis (TB) \cite{Akhter2011}, Metric Projections (MP) \cite{Paladini2012}, VA \cite{Garg2013} and Dense Spatio-Temporal Approach (DSTA) \cite{Dai2017} are replicated from \cite{Dai2017}, and the numbers for Coherent Depth Fields (CDF) \cite{Golyanik2017}, Grassmannian Manifold (GM) \cite{Kumar2018} and Jumping Manifolds (JM) \cite{Kumar2019} are taken from the original papers. We compute RMSE for SMSR \cite{Ansari2017} as the authors reported another metric. 

Our approach is ranked fourth out of nine, and the gap between the most accurate methods is far less than $10^{-2}$, which does not allow to generalise this result  with confidence. 
Our RMSE is remarkably close to the currently most accurate GM/JM methods on these sequences, even though we design D-CMDR based on simpler principles. 
By submitting our results to the NRSfM challenge \cite{Jensen2018}, we additionally compare the proposed D-CMDR against more than fifteen methods, including TB, MP and SMSR on five evaluation scenarios. DSTA, CDF, GM and JM are not compared on the NRSfM challenge yet. 
As this dataset targets sparse and semi-dense reconstructions, we disable $\bfE_{\text{reg.}}$ and achieve the overall RMSE of $50.19mm$ outperforming multiple recent methods 
\cite{Dai2014, Kong_2016, Chhatkuli2016, Lee2016} and coming close to \cite{DelBue2010} ($48.79mm$). 
For the \textit{tricky} camera path, we obtain the RMSE across all sequences of $46.74mm$, which is among the best four results \cite{Jensen2018}. 
The most accurate camera trajectory for us is \textit{zigzag} with RMSE of $36.69mm$ (ranked average across all methods). 

\vspace{7pt} 

\subsection{Self- and Cross-Convergence Tests}\label{subs:main_experiment} 

The results of self- and cross-convergence tests with missing data and perturbed tracks are summarised in Fig.~\ref{fig:STATISTICS_2}. 
We ascertain that --- due to the decoupling property of shapes and poses  --- DSP can be retrieved on a sequence with %
different shape poses compared to poses of the incoming measurements in the online mode, with virtually no influence on the reconstruction accuracy. 
\\
\textbf{Missing Data.} The amount of missing data is varied in the range $[0; 99]\%$. We observe that at $30\%$, RMSE largely stabilises,  
and QE is very stable even with up to $75\%$ of missing data for three cases out of four. This shows that much fewer points are often 
sufficient to recover the camera pose. 
In the cross-convergence test for \textit{seq.}~$A$, a $50\%$ threshold is identifiable for two DSP generation methods (VA and SMSR). 
After surpassing the threshold, the standard deviation of QE gradually increases, up to the exception of D-CMDR. 
\begin{figure}[t!] 
\centering 
  \includegraphics[width=.99\linewidth]{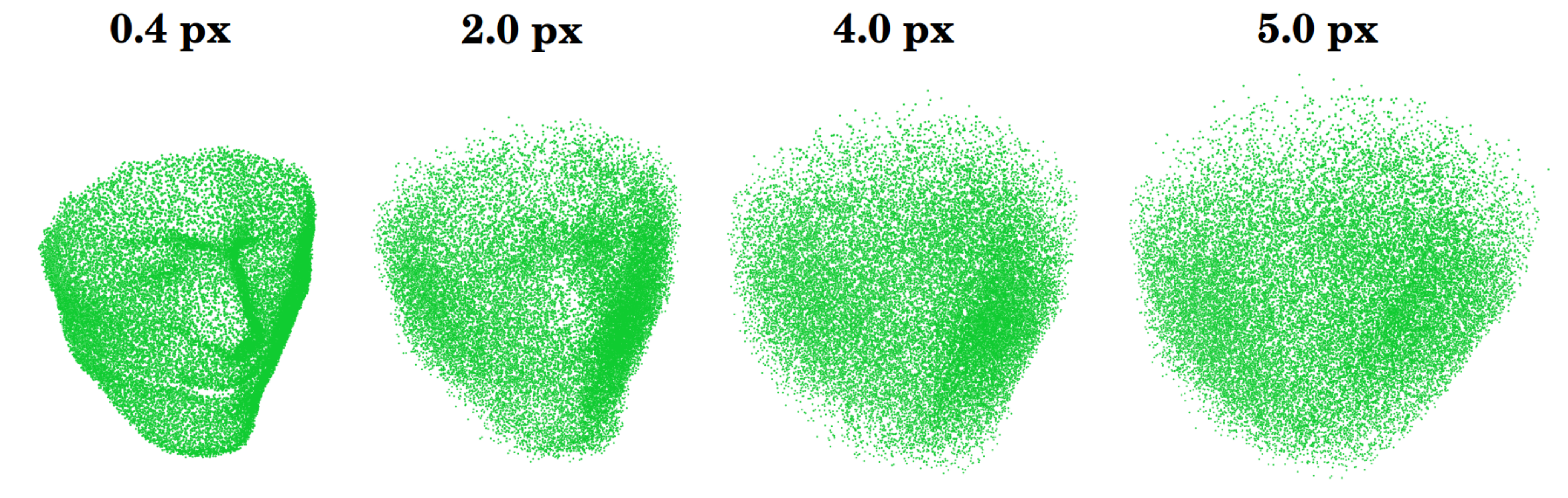} 
  \caption{ Reconstruction results of SMSR \cite{Ansari2017} on the perturbed point tracks (four different perturbation magnitudes), for frame $11$ of \textit{seq.~A}. 
  } 
  \label{fig:PERTURBED_EXAMPLE} 
\end{figure} 
In the latter case, QE is stable across the range of missing data patterns up to $90\%$. \\
\textbf{Perturbed Tracks.} In the case of the perturbed data, DSPR is stable and accurate in the whole tested range of $[0.1; 15]$ pixels  
of uniform perturbance per pixel. Across all experiments and test cases, 
RMSE is kept on the same level of accuracy and is nearly uninfluenced by the perturbations. 
On the contrary, QE is slightly affected by the increasing perturbation amplitude. 
Still, there is no observable qualitative difference in the estimation of the camera poses. 

Altogether, this is a notable result. %
Fig.~\ref{fig:PERTURBED_EXAMPLE} congregates selected outcomes of SMSR on perturbed point tracks of \textit{seq.~A} 
arranged in ascending order of deterioration. 
As the perturbation magnitude increases, 
the point scattering effects become more distinct. 
Already at $3$-$4$ pixels, the structure is barely recognisable. 
Next, the appearance 
obtained on the tracks with missing data is reasonable but contains missing entries. 
Suddenly, with $23$-$25\%$ of missing entries, no meaningful structure can be 
reconstructed by SMSR. %

Table~\ref{table:comparisons_other_methods} summarises the metrics. 
Error patterns of the plain D-CMDR and SMSR are comparable. 
In contrast, DSPR operates on the tracks with amounts of missing data exceeding $25\%$. 
Even though the accuracy drops by the factor of $2$-$3$, the structure remains recognisable, and the accuracy of camera pose estimation is only marginally affected. 

\begin{figure}[t!] 
\centering 
  \includegraphics[width=.99\linewidth]{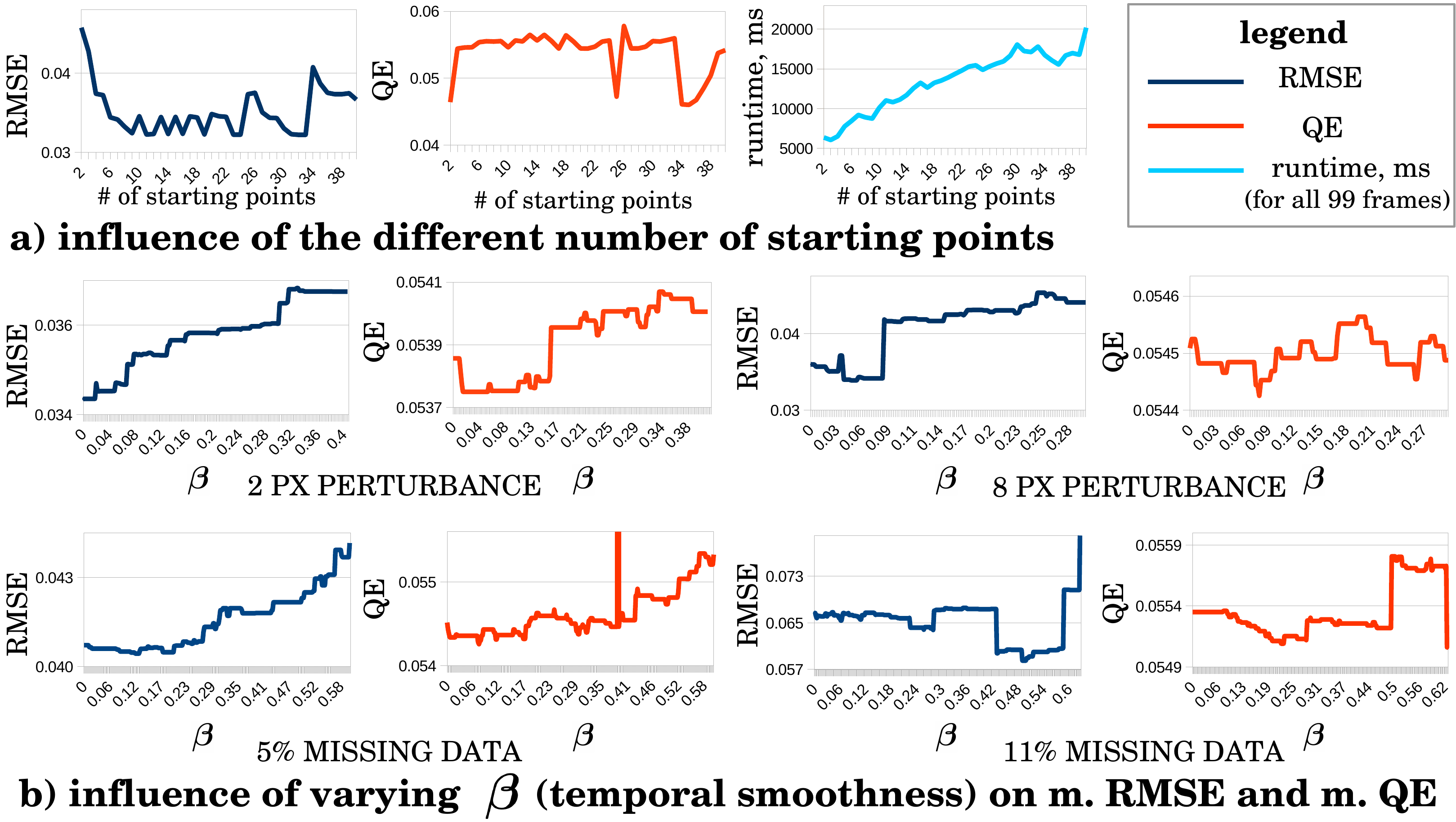} 
  \caption{ %
  Results of the experiments with MSGD parameters: (a) the influence of the different number of starting points is evaluated on 
  the measurements without noise; (b) the influence of $\beta$ is evaluated on perturbed tracks and tracks with missing data 
  with $20$ starting points. In both cases, the shapes of \textit{seq.}~$B$ are taken as a DSP and the clean tracks of 
  \textit{seq.}~$A$ are taken as the incoming dense point tracks. Mind the scaling of the $y$-axis. 
  } 
  \label{fig:SGD_GRAPHS} 
\end{figure}

\subsection{Influence of the MSGD Parameters}\label{ssec:SGD_parameters} 

The goal of this test is to examine the influence of the number of MSGD seeds, and verify that the temporal smoothness term affects reconstructions while optimising \eqref{eq:DSPR_SEARCH}. % 
Therefore, we fix $\alpha$ and vary $\beta$ (in the range $[0.0; 0.63]$ with the step $2 \cdot 10^{-3}$) under a different number of MSGD starting points (in the range $[2; 40]$), see Fig.~\ref{fig:SGD_GRAPHS}. 
\\
\textbf{Varying Number of MSGD Starting Points.} 
As expected, the runtime increases with the increasing number of seeds, and the dependency is close to a linear, see Fig.~\ref{fig:SGD_GRAPHS}-(a). Starting from $6$ seconds for two points, the runtime increases to $20$ seconds for $40$ points (for all $99$ frames). 
For $10$ and $25$ starting points, RMSE is the smallest. 
In this region, we observe oscillations of the growing period and amplitude caused by regular shifts of the starting points and different convergence due to different camera poses. 
M.~QE, on the contrary, does not correlate with the pattern of RMSE much and keeps at \textit{ca.}~$0.055$. 
The latter phenomenon stems from the decoupled nature of the geometry and camera poses. \\ 
\textbf{Varying $\boldsymbol{\beta}$.} 
Next, we vary $\beta$ under four different types of noise --- $2$ and $8$ pixels of uniform perturbances and $5\%$ and $11\%$ of missing entries (Fig.~\ref{fig:SGD_GRAPHS}-(b)). 
With small disturbances ($2$ pixels perturbance and $5\%$ of missing data), RMSE and QE vary slightly. 
The lowest errors are reached with small $\beta$. 
By and large, the errors are smaller for the case of smaller disturbances. 
For $8$ pixels perturbance and $11\%$ of missing data, 
the optimal metrics are achieved with a larger $\beta$ ($\beta \approx 0.05$ for $8$ pixels perturbance and $\beta \approx 0.5$ for $11\%$ of missing data) suggesting that the shape smoothness term is more effective for more noisy point tracks. 

\subsection{Joint Evaluation of Flow and DSPR}\label{ssec:joint_evaluation}

Joint evaluation of input flow fields and NRSfM considers the influence of the dense correspondence tracking on the reconstructions. 
Even though still not being widespread in the NRSfM literature, it is highly relevant for practical scenarios. 
We perform a joint evaluation of DSP generation, the influence of optical flow and DSPR on the adapted \textit{actor mocap} sequence \cite{Valgaerts2012} of $100$ frames with $3.5 \cdot 10^4$ points in each shape. 
It contains ground truth geometry, camera poses, corresponding rendered images, a reference image with the face segmentation mask and ground truth multi-frame optical flow (MFOF), \textit{i.e.,} a series of optical flows between the reference frame and every other frame in the sequence. 
In our modification, we rotate the ground truth surfaces and project them onto an image plane by ray tracing to render the images and the mask. 
The ground truth MFOF is obtained as the distances between the projections of the corresponding points in the image plane.

\begin{figure}[t!] 
\centering 
  \includegraphics[width=.99\linewidth]{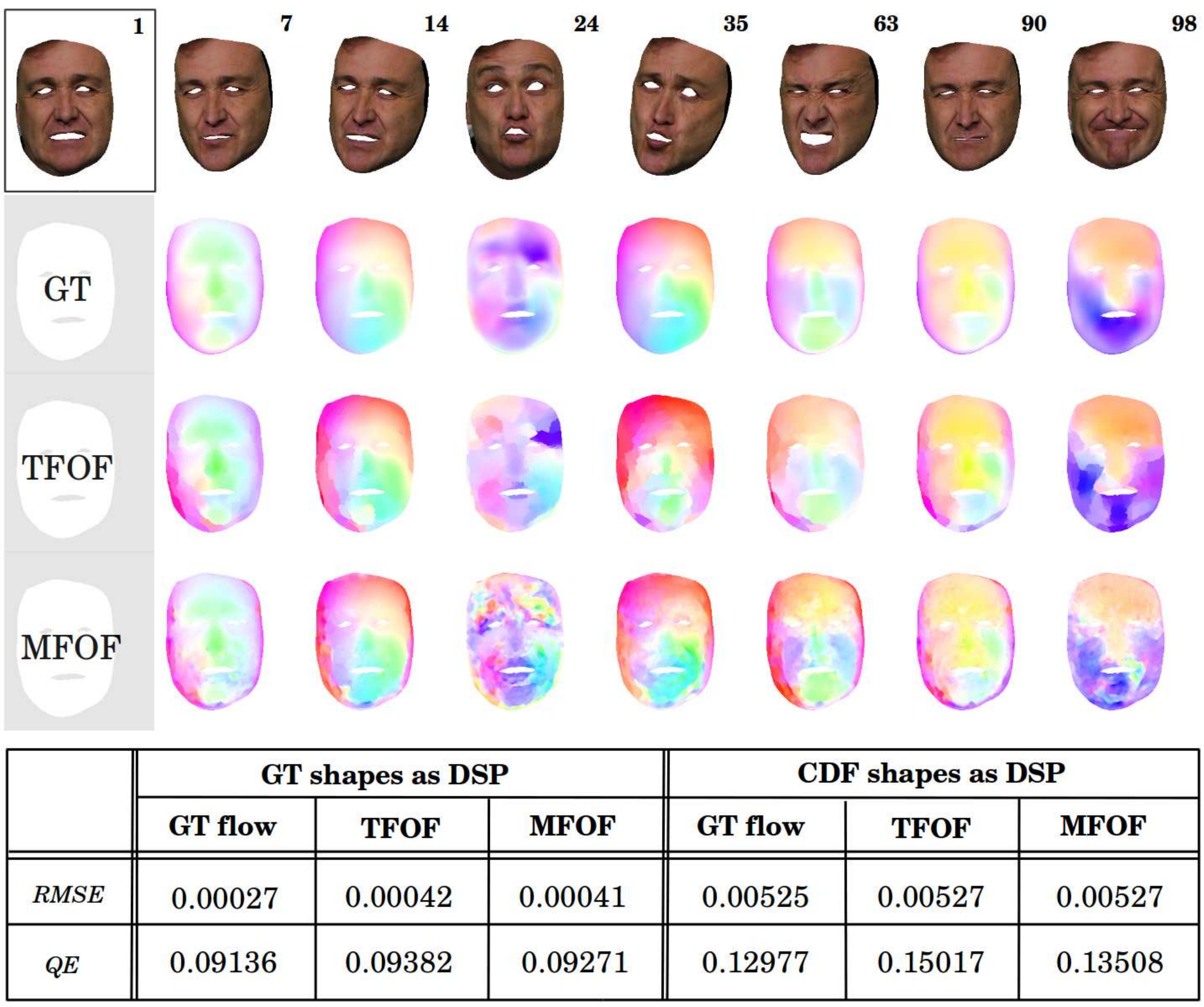} 
  \caption{ Exemplary frames from the adapted \textit{actor mocap} sequence (first row), corresponding ground truth dense flow (second row), flow obtained by the method of Sun \textit{et al.}~\cite{Sun2010} (third row) and MFOF \cite{Garg2013} (fourth row). The table underneath lists errors for all evaluated combinations. } 
  \label{fig:actor_mocap} 
\end{figure}

In addition to the ground truth MFOF, we compute dense correspondences by the method of Sun \etal \cite{Sun2010} in the pairwise manner, 
as well as global MFOF with point trajectory regularisation over the whole batch \cite{Garg2013flag}. The \textit{average endpoint error} (AEPE) of two-frame optical flow (TFOF) and MFOF amount to 1.218 and 1.123, respectively. Next, we evaluate DSPR with the ground truth flow, TFOF and MFOF while using as DSP either ground truth geometry or shapes obtained by CDF \cite{Golyanik2017} on the MFOF point tracks. 
In both cases, DSP contains $65$ states after the compression. 
Fig.~\ref{fig:actor_mocap} shows exemplary images and different types of flow fields, while the associated table summarises the results. 
We see that the errors achieved with MFOF are slightly and consistently more accurate than those obtained with TFOF. 
Still, the TFOF errors do not worsen much attesting that DSPR tolerates less reliable and noisy point tracks. 

\begin{figure}[t!] 
\centering 
  \includegraphics[width=.99\linewidth]{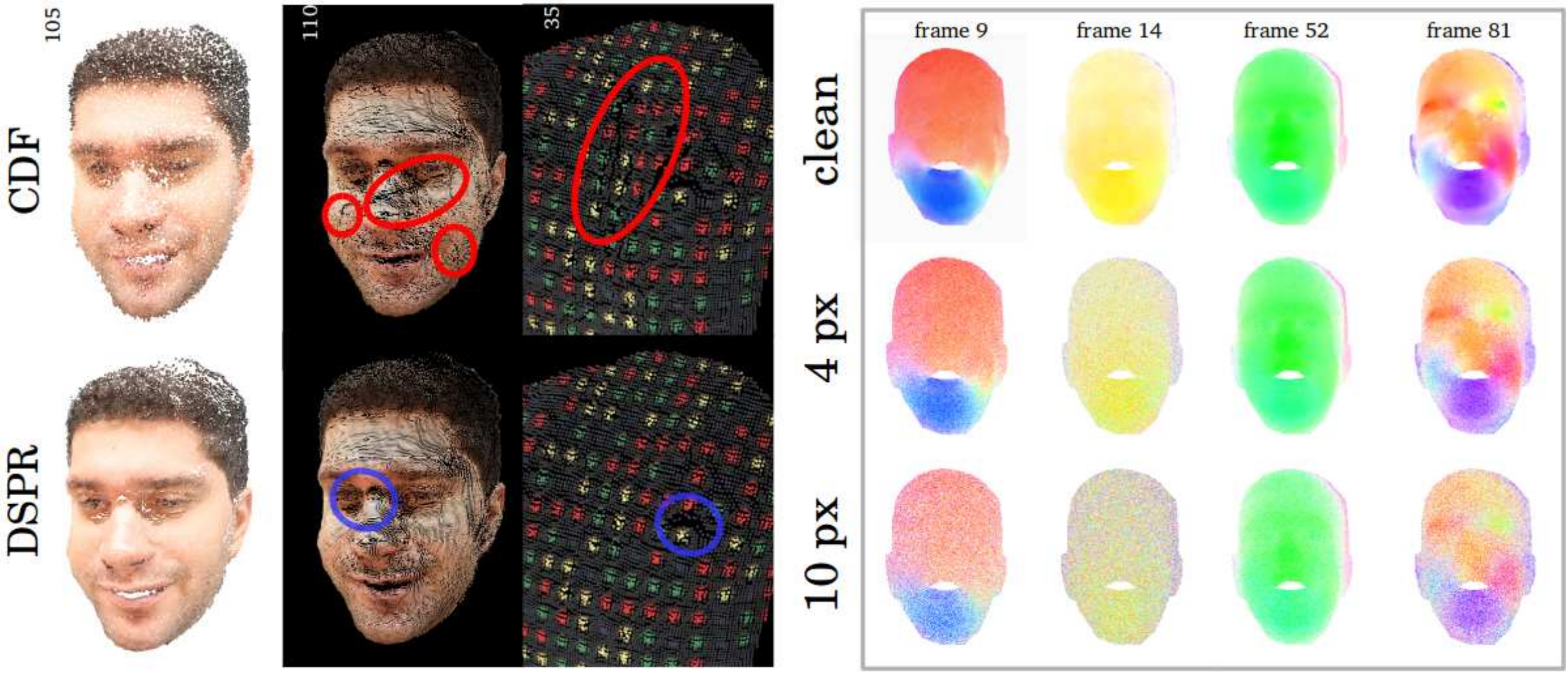} %
  \caption{ Exemplary reconstructions of CDF \cite{Golyanik2017} and DSPR on point tracks with $10$ pixels of perturbation magnitude 
  (left column) and the comparison of the compressed states (the second and third columns). 
  Compression artefacts highlighted in red are more pronounced for CDF, even though it achieves $~2.34$-$2.65$ times smaller compression ratio. 
  The blue circles emphasise artefacts due to the tracking. %
  The examples of clean and noisy point tracks with perturbances of $4$ and $10$ pixels are on the right. 
  } 
  \label{fig:FACES_COMPARISON} 
\end{figure}

\subsection{Experiments with Real Data and Applications}\label{ssec:experiments_with_real_data}

  We perform tests with real \textit{face} \cite{Garg2013}, \textit{back} \cite{Russel2011}, \textit{liver} \cite{Mountney2010} 
  and two \textit{heart} \cite{Stoyanov2005, Stoyanov2012} sequences. 
  Apart from the monocular non-rigid reconstruction, several other modes of operation are conceivable for DSPR. First, if we rerun DSPR on 
  the point tracks which are used to compute DSP, we obtain a compressed version of the reconstructions. 
  With the increasing density and the number of views, the space required for storage of a dynamic reconstruction grows fast. 
  Especially in embedded and mobile devices, limits on the data bandwidth can become a bottleneck. Hence, compression of dynamic reconstructions is of high practical relevance. 
  In the compression mode, we need to save a DSP, a shape prior identifier and a camera pose for every frame. 
  This adds up to $12$ bytes for camera pose in the axis-angle representation and one-two bytes for the shape prior identifier. 
  Second, we are free to mix the sources of the DSP and incoming measurements. 
  By computing correspondences between a reference frame of one sequence and frames observing a similar scene from another sequence, we can reenact 3D deformation states as if they were another scene. \\ %
\textbf{DSPR for Shape Compression.} We compare DSPR and CDF \cite{Golyanik2017} --- which is explicitly designed for compressible representations --- for shape compression. We 
\begin{table}[!t]
  \footnotesize
  \begin{center}
   \begin{tabular}{|c|c|c|c|c|c|} \hline 
   \multirow{2}*{$\;\boldsymbol{\mu}$} 	&  \multirow{2}*{\textbf{MSGD seeds}} 	& \multicolumn{2}{c|}{\textbf{\textit{face} \cite{Garg2013}}} 			& \multicolumn{2}{c|}{\textbf{\textit{back} \cite{Russel2011}}} 	\\\cline{3-6} 
						&					     	& \textbf{$\;\;|$DSP$|\;\;$} 	& \textbf{$\;\;\;C\;\;\;$} 	& \textbf{$\;\;|$DSP$|\;\;$} 	& \textbf{$\;\;\;C\;\;\;$}  	\\\hline \hline 				
    1.5						&	20					& 73				& 1.64				& 67			& 2.23					\\\hline 	
    2.5						&	20					& 59				& 2.03				& 57			& 2.63					\\\hline 
    5						&	20					& 42				& 2.85				& 40			& 3.75					\\\hline 
    10.0					&	12					& 25				& 4.8				& 25			& 6					\\\hline 
    20.0					&	8					& 15				& 8				& 16			& 9.375					\\\hline 
    30.0					&	5					& 10				& 12				& 11			& 13.63					\\\hline 
    40.0					&	4				 	& 8				& 15				& 9			& 16.6					\\\hline 
   \end{tabular} 
  \end{center} 
  \vspace{-10pt}
  \caption{ %
  The summary of the experiment for the compression of dynamic reconstructions with the achieved compression ratios. 
  } 
  \vspace{-7pt}
   \label{table:compression}
\end{table} 
\begin{figure*}[t!] 
\centering 
  \includegraphics[width=1.01\linewidth]{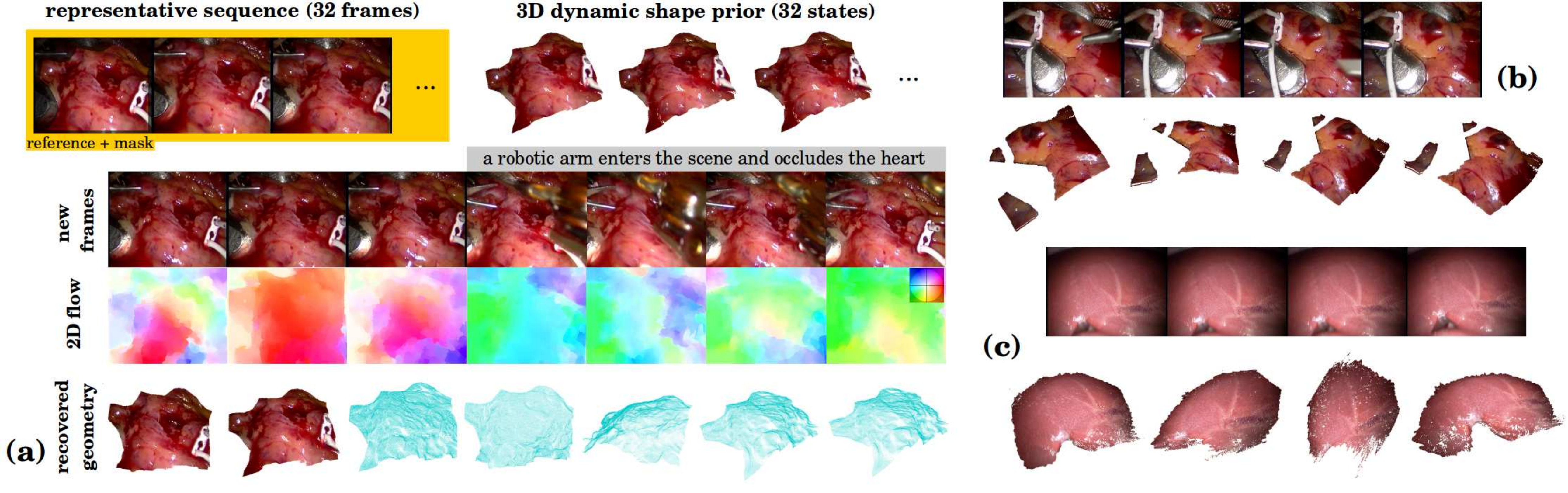} 
  \vspace{-15pt}
\caption{ 
Application of DSPR in endoscopic scenarios with pronounced reoccurring deformations. 
(a): The first \textit{heart} sequence \cite{Stoyanov2005}. 
The representative sequence and exemplary DSP states are shown in the top row. 
The new incoming frames and calculated flow fields visualised with Middlebury colour scheme \cite{Baker2011} are given underneath. 
Our reconstructions from different perspectives are displayed in the bottom row. 
(b), (c): Exemplary frames and our reconstructions of another \textit{heart} \cite{Stoyanov2005} and the \textit{liver} sequences \cite{Mountney2010}, respectively. 
Best viewed in colour. Furthermore, see our supplementary video. 
} 
\label{fig:HEART_SEQUENCE} 
\end{figure*} 
use MFOF \cite{Garg2013flag} point tracks of \textit{face} and \textit{back} sequences, and extra prepare perturbed measurements of \textit{face}. For the latter, CDF achieves compression ratio $C = 7.0$ on clean tracks. On the noisy tracks with the perturbation magnitudes of $4$ and $10$ pixels, its $C$ decays to $3.0$ and $1.582$, respectively ($\epsilon = 1.6 \cdot 10^{-3}$). 
DSPR reaches $C = 8.0$ for $\mu = 20.0$ under $10$ pixels of perturbations. %
If DSP is computed on clean reconstructions, the compression ratio is only weakly affected by the noise in point tracks, and only slight qualitative differences can be noticed (see the supplementary video). 
On the \textit{back}, CDF achieves $C = 4.0$ ($\epsilon = 8 \cdot 10^{-4}$) and DSPR converges at $C = 9.375$ with $\mu = 20.0$. %
Recall that for DSPR, the longer an image sequence is, the higher are the compression ratios. 
The compression quality depends on how accurate the representative sequence for DSP generation reflects the shape space in the interactive mode. 
Fig.~\ref{fig:FACES_COMPARISON} compares the reconstructions obtained by CDF and DSRP. 
The left column shows the resulting states 
on noisy tracks ($10$ pixels of perturbation magnitude). 
As we see, especially with high compression ratios, CDF causes noticeable compression artefacts. 
For DSPR, Table~\ref{table:compression} reports all combinations of the tested thresholds $\mu$, corresponding DSP cardinalities, the number of MSGD seeds 
and the attained compression ratios for the \textit{face} and \textit{back} sequences. \\ 
\textbf{DSPR in Endoscopy}. Scenarios with temporally-disjoint rigidity often occur in the endoscopy. 
Examples are a human heart undergoing a series of repetitive contractions or a liver with periodic respiratory deformations. 
We test the proposed DSPR on two \textit{heart} sequences from Stoyanov \textit{et al.}~\cite{Stoyanov2005, Stoyanov2012} and a \textit{liver} 
sequence from \cite{Mountney2010}. 

The first \textit{heart} sequence contains $1573$ frames in total\footnote{shorter parts of this sequence are often used in NRSfM \cite{Agudo_Moreno_Noguer_2015, Ansari2017, Dai2017, Kumar2018}}. For the DSP reconstruction, we choose $32$ unoccluded frames --- this duration corresponds to one complete cardiac cycle --- and compute MFOF \cite{Taetz2016}. 
Next, we reconstruct a DSP with $32$ states and $68k$ points per state, see Fig.~\ref{fig:HEART_SEQUENCE}-(a). 
The geometries during the diastole (refilling) and the systole (contraction) are all different, and, hence, we \textit{do not} perform state compression. 
Next, we compute TFOF \cite{Sun2010} between the reference and every remaining frame, and execute DSPR achieving five frames per seconds 
on our system (${\sim}4.1$Mbps for the incoming optical flow). %
The reconstruction follows the cardiac cycle, \textit{even if the robotic arm partially occludes the heart}. 
Following similar steps, we reconstruct the second \textit{heart} sequence ($899$ frames, $55k$ points per surface), 
and the \textit{liver} ($250$ frames, $54.5k$ points per surface). 
The \textit{heart} sequence in Fig.~\ref{fig:HEART_SEQUENCE}-(b) also has occlusions, and the reconstruction distinctly reflects the cardiac cycle. 
The \textit{liver} in Fig.~\ref{fig:HEART_SEQUENCE}-(c) contains large displacements in the point tracks. 
See our video with dynamic visualisations. 

\subsection{Discussion} 

We see that the substitution of deformation weights in classic low-rank NRSfM by a selection mechanism for each DSP element results in a fast and robust dense sequential NRSfM. 
We also witness that the proposed optimisation procedure is very quick, possibly faster than most of the NRSfM algorithms in the literature. 
In the online mode, DSPR achieves multiple frames per second, and the throughput can be further increased by parallelisation. 

All in all, we match the state-of-the-art performance while drastically improving the robustness on deteriorated point tracks --- in reality, data is often far from perfect. 
Last but not least, the compression evaluation is also rarely seen in NRSfM but is of practical relevance. 
\section{Conclusions and Future Directions} 
\label{sec:conclusion} 

We introduce a new hybrid NRSfM method relying on temporally-disjoint rigidity effects. 
In the first step of our DSPR approach, we reconstruct a representative set of views and generate DSP. 
Next, for new incoming dense point tracks, we solve a light-weight optimisation problem with a zero-norm which selects the closest shape from DSP while positioning it as observed in the measurements. 
The robustness to inaccurate point tracks, the possibility to use faster and less accurate dense flow fields, the highest compression ratios and the suitability of the proposed technique for medical applications with repetitive deformations significantly broaden the scope of modern NRSfM, especially when handling real data. 
We show experimentally that DSPR successfully bridges the gap between the accuracy of dense correspondences and reconstructions, and we believe that it can have a high practical impact. 
Furthermore, our light-weight dense incremental NRSfM can enable various new applications in augmented reality.

There are multiple future work directions. 
Though not explicitly tested, the sparse setting can also be investigated, because runtime of dense NRSfM has always been more of an issue. 
DSPR can be deployed on a low-power consumption device such as augmented reality glasses for applications involving deformable objects. 
Moreover, DSP signatures are worth trying for object class recognition. 
%

%\section*{Appendix} 

%{\small 
%Our supplementary material contains a video with experimental results, the adapted \textit{actor mocap} dataset for the joint evaluation of flow and NRSfM, 
%as well as a detailed description of the results obtained on the NRSfM challenge \cite{Jensen2018}. 
%} 

\appendix 

\section{Appendix}\label{app:appendix} 

\renewcommand{\thefigure}{\Roman{figure}} 
\setcounter{figure}{0} 

\renewcommand{\thetable}{\Roman{table}} 
\setcounter{table}{0} 

\renewcommand{\theequation}{\roman{equation}} 
\setcounter{equation}{0}

This section provides more insights about the DSPR approach including a remark on rotational ambiguities (Sec.~\ref{app_ssec_1}), 
convergence patterns (Sec.~\ref{app_ssec_2}) as well as more results on real data and NRSfM challenge dataset (Secs.~\ref{app_ssec_3}, \ref{app_ssec_4}). 
% 
% Dynamic Shape Prior Reconstruction (DSPR) approach for monocular non-rigid 3D reconstruction. 
% 
In addition, our paper is supplemented by a four and a half minutes long video\footnote{\url{http://gvv.mpi-inf.mpg.de/projects/DSPR/}}. 
% 
% showcasing the results indicated in the main paper 

% I. Rotational ambiguities in orthographic NRSfM and their influence in DSPR 

\subsection{On Rotational Ambiguities in DSPR}\label{app_ssec_1} % NRSfM and 

\begin{figure}[t!] 
\centering 
  \includegraphics[width=.99\linewidth]{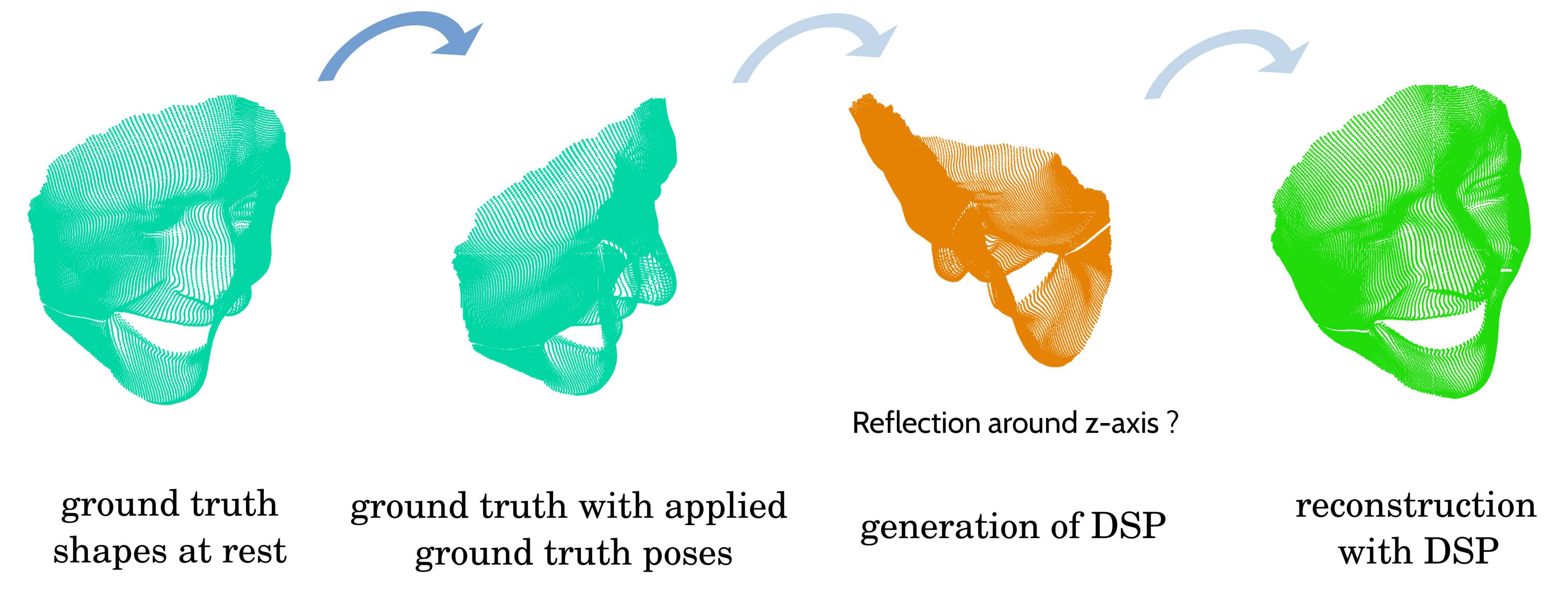} 
  \caption{ 
  For the explanation of the rotational and reflectional ambiguities arising in DSPR. 
  Ground truth is provided in an arbitrary reference frame (left), and ground truth rotations are applied to 
  the ground truth shapes at rest in this reference frame (middle-left). 
  DSP is reconstructed from dense point tracks in a different and arbitrary reference frame (middle-right). 
  DSP reconstructions can be additionally reflected w.r.t.~ground truth. 
  Reconstruction with DSP fetches the most suitable shape and places it in the observed pose (right). 
  Dark and light blue curved arrows indicate that the rotations between the shapes are known and unknown, respectively. 
  } 
  \label{fig:ROT_AMBIGUITY} 
\end{figure}

When computing DSP, we have to take into account rotational and reflectional ambiguities arising in orthographic NRSfM. % have to be considered. 
Rotational ambiguity refers to the arbitrariness of the global reference frame of NRSfM reconstructions. 
While the reconstructions are correctly positioned relative to each other, the information about their global orientation is not contained in the measurements. 
% For DSP the only thing which matters 
The implication for DSPR is that all $\bfD_i$ have to be positioned in one arbitrary reference frame which the same for all $\bfD_i$. 
Reflection of the reconstruction around the $z$-axis always occurs under rigid initialisation with an orthographic camera since singular value decomposition is ambiguous in the sign. 
It propagates to DSPR with no consequence for the method from the numerical point of view. 
Fig.~\ref{fig:ROT_AMBIGUITY} exemplifies the rotational and reflectional ambiguities of NRSfM as occurred in the experiment with the synthetic face sequences. 
% 

% II. Convergence pattern visualisation 

\subsection{Convergence Patterns}\label{app_ssec_2} % in Convergence Tests 

\begin{figure*}[t!] 
\centering 
  \includegraphics[width=.94\linewidth]{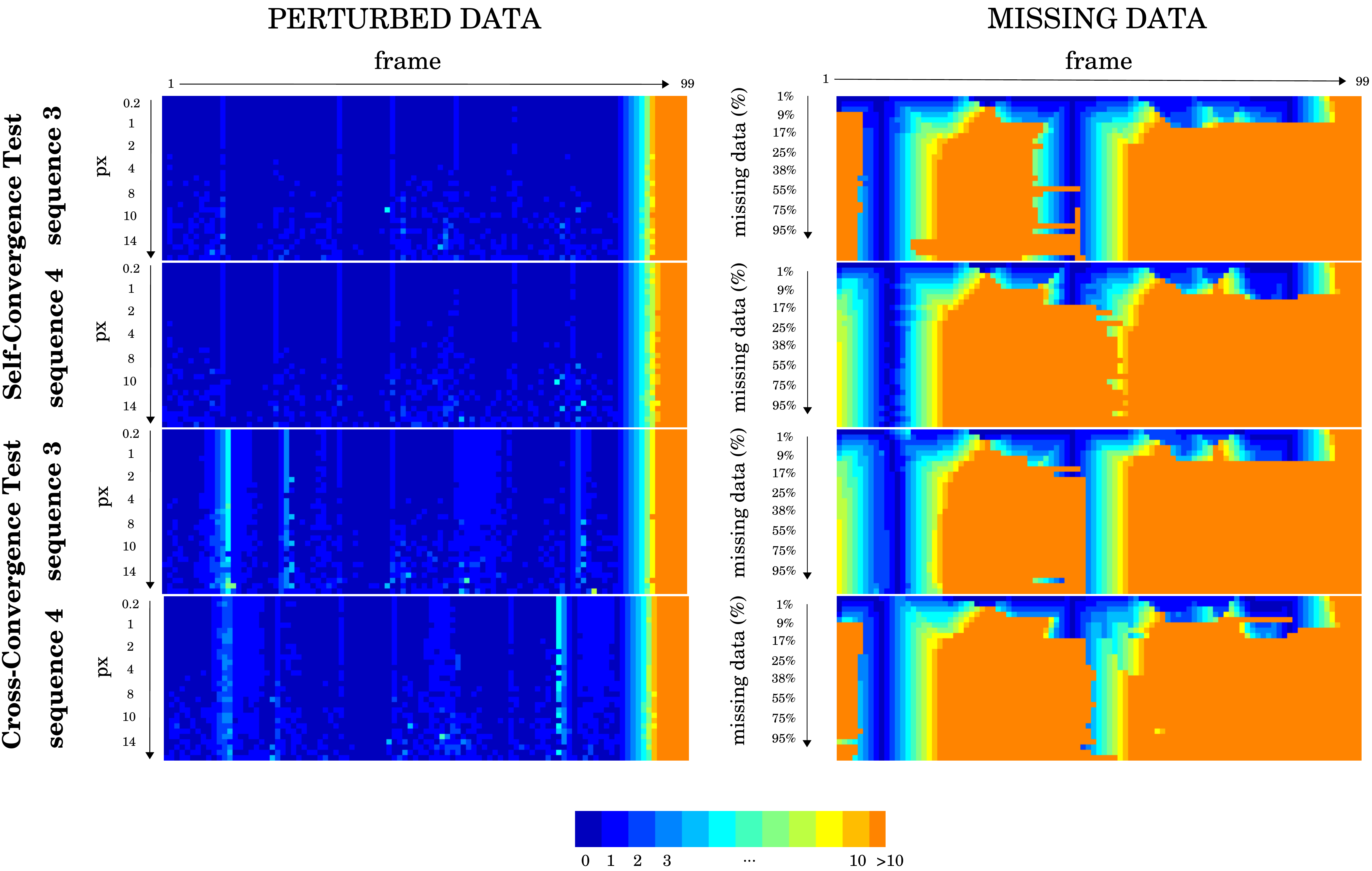} 
\caption{ Convergence patterns observed in the self- and cross-convergence tests. Every square stands for the $\eta$ shape distance for 
a given frame ($x$-axis) and type of the noise ($y$-axis, columnwise depending on the type of noise). The colour coding scheme for $\eta$ is provided 
beneath. } 
\label{fig:convergence_pattern} 
\end{figure*}

A \textit{convergence pattern} in DSPR refers to the sequence of states chosen from DSP for every new incoming measurement. 
In the self- and cross-convergence tests (Sec.~\ref{subs:main_experiment}), it is possible to analyse convergence patterns quantitatively because the ground truth state identifiers are known.  As a quantitative metric, we use the absolute distance from the chosen DSP state and the ground truth state % which was used 
for inducing the measurements denoted by $\eta$. 
Recall that two face sequences from \cite{Garg2013} were used in the self- and cross-convergence tests. They contain $99$ frames obtained from 
ten basis facial expressions by interpolation and differ in the applied series of rotations. The sequences are originally referred to % to them 
as \textit{sequence $3$} and \textit{sequence $4$}. 
Convergence pattern is an auxiliary metric as it does not take into account that DSP states can be similar or both well 
explain 2D observations. Consequently, even if $\eta$ is large, $e_{3D}$ can be small, and, conversely, large $\eta$ can imply incorrect 
convergence and a high $e_{3D}$. 

In Fig.~\ref{fig:convergence_pattern}, convergence patterns for all self- and cross-convergence experiments are visualised. 
We observe that the convergence pattern of the self-convergence test mostly contains small $\eta$, except for the last ten shapes. 
The reason is that the last ten shapes are similar to other shapes of the interpolated sequence (\textit{e.g.,} shape $94$ is similar 
to shape $71$), DSPR chooses a state with a higher $\eta$ as a solution, and both states lead to low $e_{3D}$. 

The convergence pattern of the cross-convergence test is slightly different, as the structure is observed in different poses. 
Larger $\eta$ at frame $15$ and in the vicinity show that the convergence of DSPR is dependent on the accuracy of camera pose estimation. 
Moreover, note the differences in the convergence patterns due to either increasing perturbation magnitude or the missing data ratio. 
Some shapes are more sensitive to the disturbing effects compared to the others, which can be explained by a decaying resolvability, 
\textit{i.e.,} the ability of the method to distinguish between the shapes. 
% the to decide between shapes explaining the observations equally well because of disturbing effects. 
% Resolution problem 

% III. Shape compression and monocular reenactment (2D-3D expression transfer) 

\subsection{More Results on Real Data}\label{app_ssec_3}

\begin{figure}[t!] 
\centering 
  \includegraphics[width=.99\linewidth]{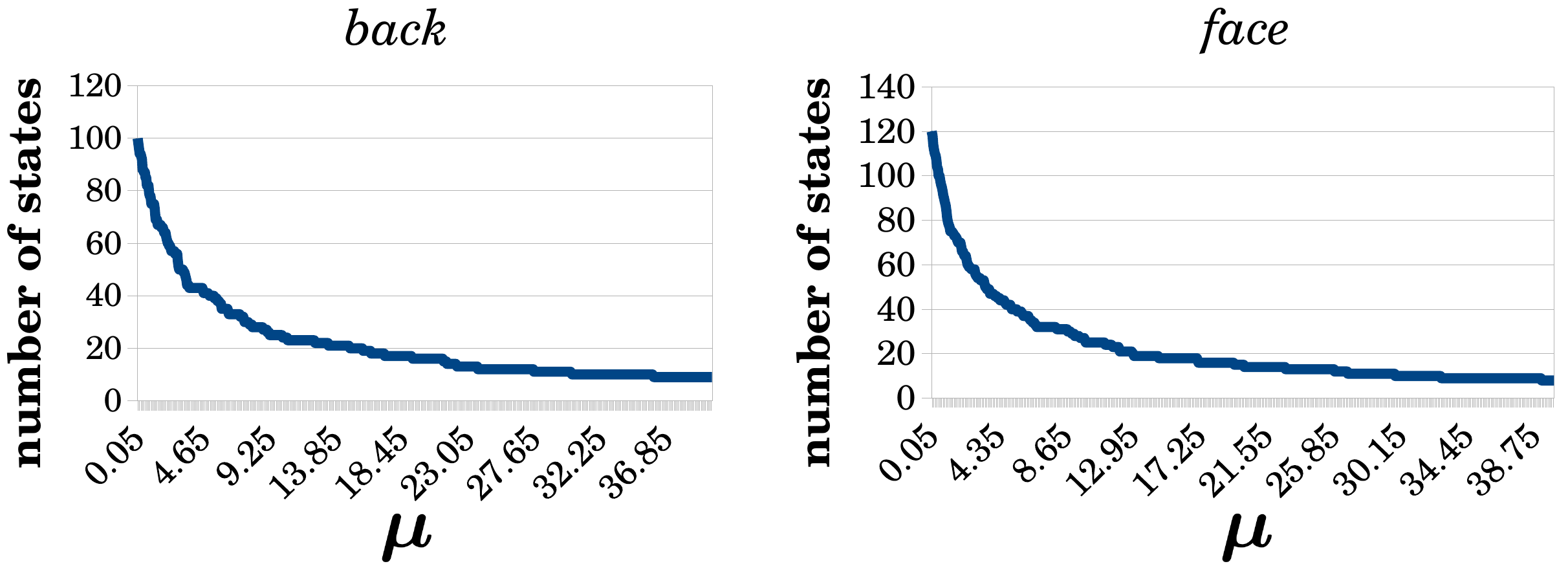} 
\caption{ DSP cardinalities as functions of the threshold $\mu$, for the \textit{face} 
\cite{Garg2013} and \textit{back} \cite{Russel2011} sequences.  } 
\label{fig:DSP_PLOTS} 
\vspace{-12pt} 
\end{figure} 

In the shape compression mode, DSPR runs on the representative image sequence which was used for DSP generation or on a longer sequence. 
An approximation of the compression ratio is obtained by division of the number of frames in the sequence by the DSP cardinality. 
The achieved compression ratio depends on the DSP cardinality, which, in turn, depends on the threshold $\mu$. For the \textit{face} 
\cite{Garg2013} and \textit{back} \cite{Russel2011} sequences, Fig.~\ref{fig:DSP_PLOTS} features the number of % dependencies 
elements in DSP as the functions of $\mu$. These step functions are monotonically decreasing, and their rate of change per interval is the measure of 
state variability of a 3D sequence. 
If a sufficient increase in $\mu$ leads to minor changes in the DSP cardinality, the 3D states are more dissimilar ($\mu > 20$ in  Fig.~\ref{fig:DSP_PLOTS}). 
If DSP cardinality drops significantly in a short interval of $\mu$ values, this indicates that many states are repeated in the sequence  ($\mu < 10$ in Fig.~\ref{fig:DSP_PLOTS}). 
% 

% The derivative of these step functions 
% The exemplary functions 

\begin{figure*}%[t!] 
\centering 
\begin{subfigure}{0.49\textwidth} 
\includegraphics[width=\textwidth]{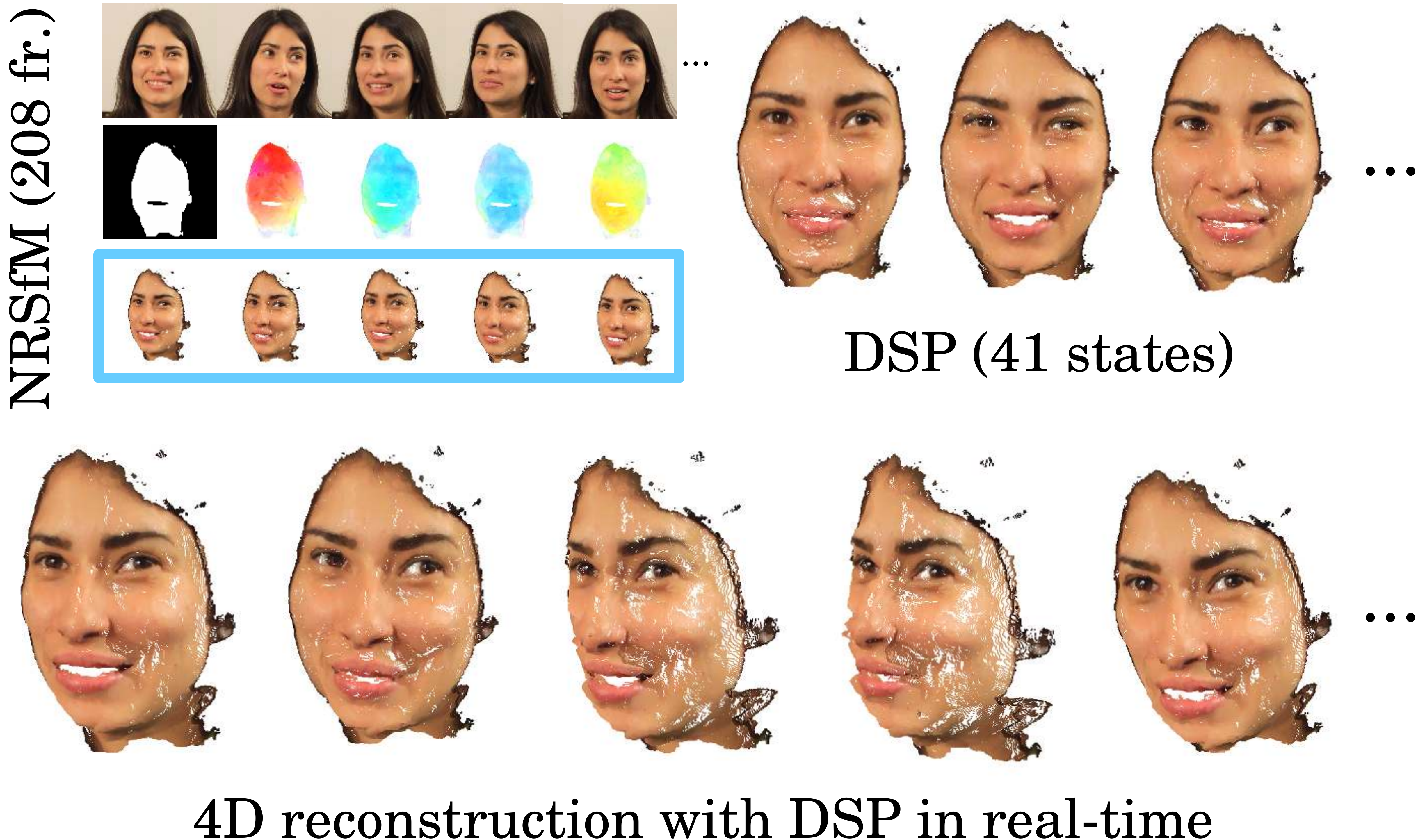} 
\caption{\footnotesize The input image sequence, dense point tracks and unrotated reconstructions for DSP estimation (top left), 
several states from the DSP (top right) and the reconstructions with DSPR of the displayed images (the bottom row). } 
\label{fig:jilliam} 
\end{subfigure}
\hspace{5pt}
\begin{subfigure}{0.49\textwidth}
  \includegraphics[width=\textwidth]{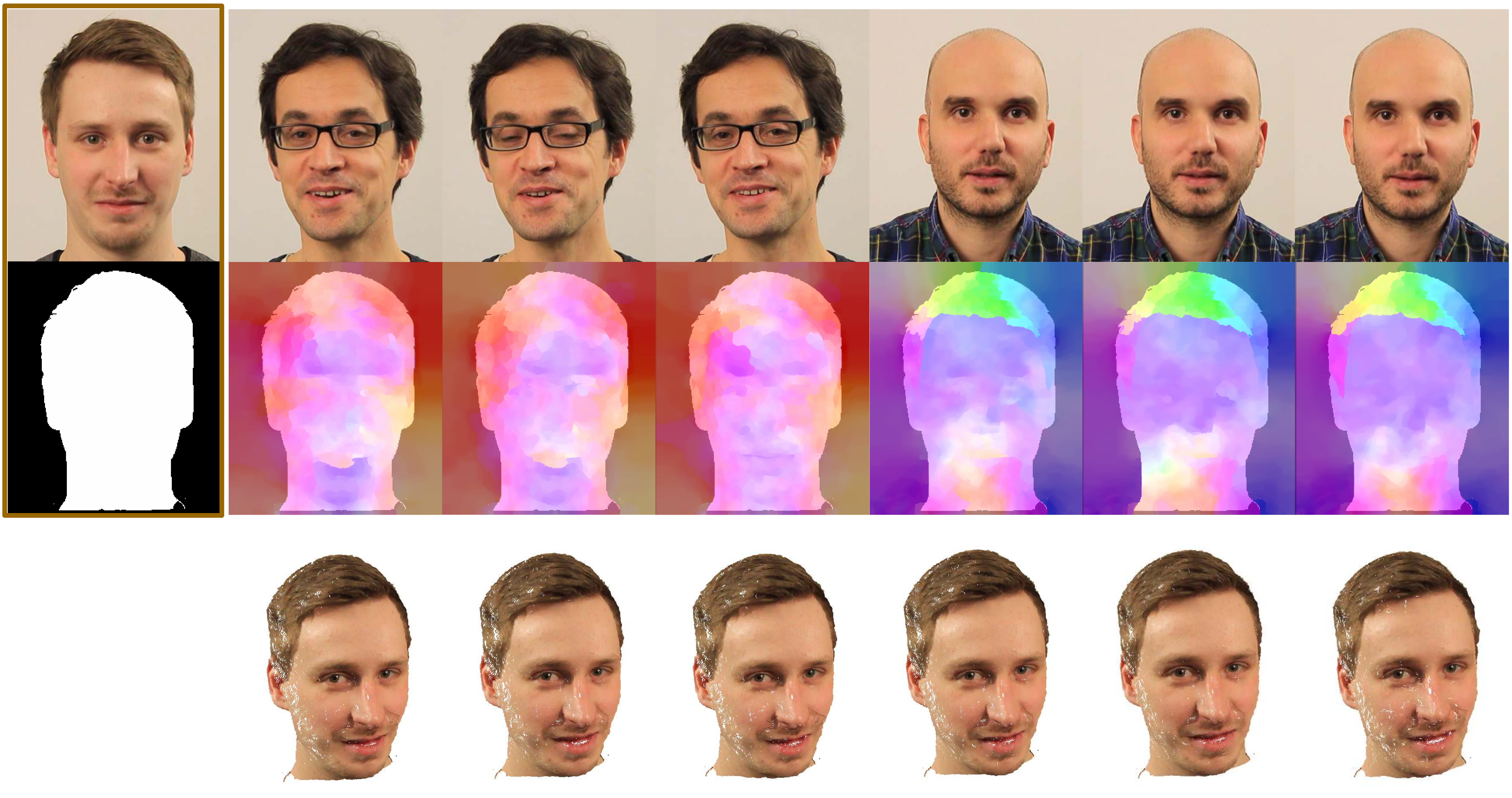} 
\caption{\footnotesize Visualisation of monocular reenactment with DSPR --- the reference frame and the DSP belong to the target sequence (on the left), 
and the measurements are obtained as warps between the reference frame and new incoming frames of the reenacting sequence by the method of 
Sun \textit{et al.} \cite{Sun2010} (top and middle rows). The bottom row shows the resulting reenactments. } 
\label{fig:reenactments} 
\end{subfigure}
\caption{Experiments with newly recorded face sequences.} 
\vspace{15pt} 
\label{fig:new_sequences_combined} 
\end{figure*}

% 
Fig.~\ref{fig:new_sequences_combined} shows results on new face sequences recorded by CANON EOS 500D. 
% 
% Fig.~\ref{fig:jilliam} shows DSPR results on a new face sequence. 
In Fig.~\ref{fig:jilliam}, the sequence contains $208$ frames of the resolution $600 \times 600$ pixels in the representative sequence. 
From the initial $208$ NRSfM reconstructions obtained on MFOF \cite{Garg2013flag} point tracks, we extract $42$ DSP states with $\mu = 10.1$. 
DSPR runs on the representative subsequence and further frames with three frames per second. 
% 
% See our supplementary video for further visualisations. 
% The results are visualised and explained with more details in the supplementary video. 
% 

%\begin{figure}%[t!] 
%\centering 
%  \includegraphics[width=.99\linewidth]{FIGURES/REAL_FACES_IV} 
%\caption{The input image sequence, dense point tracks and unrotated reconstructions for DSP estimation (top left), 
%several states from the DSP (top right) and the reconstructions with DSPR of the displayed images (the bottom row). } 
%\label{fig:jilliam} 
%\end{figure} 

%\captionsetup{subfigure]{skip=10pt}

% caption : monocular reenactment 

%\begin{figure*}[t!] 
%\centering 
%  \includegraphics[width=.99\linewidth]{FIGURES/VIDEO_SCREENSHOTS} 
%\caption{ Screenshots from our supplementary video. } 
%\label{fig:video_screenshots} 
%\end{figure*} 

Fig.~\ref{fig:reenactments} demonstrates the principle of monocular reenactment on several sequences (the resolution of the frames is $550 \times 650$). 
% when views of a similar scene (in this case, faces) are 
% When an accurate warp between different though comparable scenes can be obtained, 
Thus, DSPR can be used to animate a target scene with the states of another though similar scene (in this case, faces) observed in 2D. 
Suppose we are given a reference frame and a DSP for a sequence \textit{Alice}. We can compute dense warps between the given reference frame 
and incoming frame of the \textit{Bob} sequence and use them in the same fashion as if they would originate from the \textit{Alice} sequence. 
As a result, we obtain 3D facial expressions of \textit{Alice} as observed in 2D for \textit{Bob}. However, specific methods for intra-instance 
warps would be more accurate compared to general-purpose optical flow where the brightness constancy term is overly challenged between two frames of 
different persons, with different head poses and under different lighting conditions and textures.

% table 

\begin{table*}[!t] 
 %\footnotesize 
 \small 
  \begin{center} 
  \begin{tabular}{|c|c|c|c|c|c|c|} \hline 
  \textit{trajectory} $\downarrow$ $\backslash$ \textbf{scene} $\rightarrow$ & \textbf{articulated} 	& \textbf{balloon} 	& \textbf{paper} 	& \textbf{stretch} 	& \textbf{tearing}  	&  \makecell{\textit{mean RMSE} \\ (over camera trajectories)}     	\\ \hline\hline 
  \textit{circle}	& 75.823		&	55.664		& 81.181		& 	92.017		& 	80.873		& 77.112		\\ \hline 
  \textit{flyby}	& 44.603		&	36.293		& 56.873		& 	53.716		& 	44.139		& 47.125		\\ \hline 
  \textit{line}		& 65.914		&	35.784		& 45.796		& 	44.625		& 	29.103		& 44.244		\\ \hline 
  \textit{semi-circle}	& 46.741		& 	32.41		& 57.67			& 	55.567		& 	53.656		& 49.209		\\ \hline 
  \textit{tricky}	& 62.559		& 	34.77		& 34.509		&	58.012		& 	43.888		& 46.748		\\ \hline 
  \textit{zigzag}	& 37.544		&	31.298		& 37.017		& 	41.101		& 	36.525		& 36.697		\\ \hline\hline 
  \makecell{\textbf{mean RMSE}\\(over sequence types)}		& 55.53			& 	37.703		& 51.486		& 	57.506		& 	48.031		& \textbf{50.19}	\\ \hline 
  \end{tabular} 
  \end{center} 
  \caption{ Results of D-CMDR evaluation on the NRSfM Challenge \cite{Jensen2018}. The numbers are the mean RMSE in $mm$ for each scene and camera trajectory, including overall mean RMSE over all camera trajectories (for each scene) and all scenes (for each camera trajectory). The mean RMSE over all scenes and all camera trajectories amounts to $50.19$ $mm$.} 
  \label{tab:NRSfM_chellenge} 
\end{table*}

\subsection{NRSfM Challenge Dataset}\label{app_ssec_4} 

In this section, we provide a more detailed analysis of the results on the NRSfM Challenge dataset\footnote{\url{http://nrsfm2017.compute.dtu.dk/benchmark}}  \cite{Jensen2018} achieved by the proposed D-CMDR method for reconstruction of the representative sequence (Sec.~\ref{ssec:CMDR}). 
Even though D-CMDR targets dense reconstructions, we evaluate D-CMDR on \cite{Jensen2018} because it becomes a standard benchmark dataset for NRSfM. 
% this data set 
We disable $\bfE_{\text{reg.}}$ which assumes spatially connected regions and a regular point arrangement (or, at least, known neighbours per point). 
% the regularisation term 
In total, our energy functional contains $\bfE_{\text{fit}}, \bfE_{\text{temp}}$ and $\bfE_{\text{linking}}$. % three terms 

\begin{figure*}[t!] 
\centering 
  \includegraphics[width=.95\linewidth]{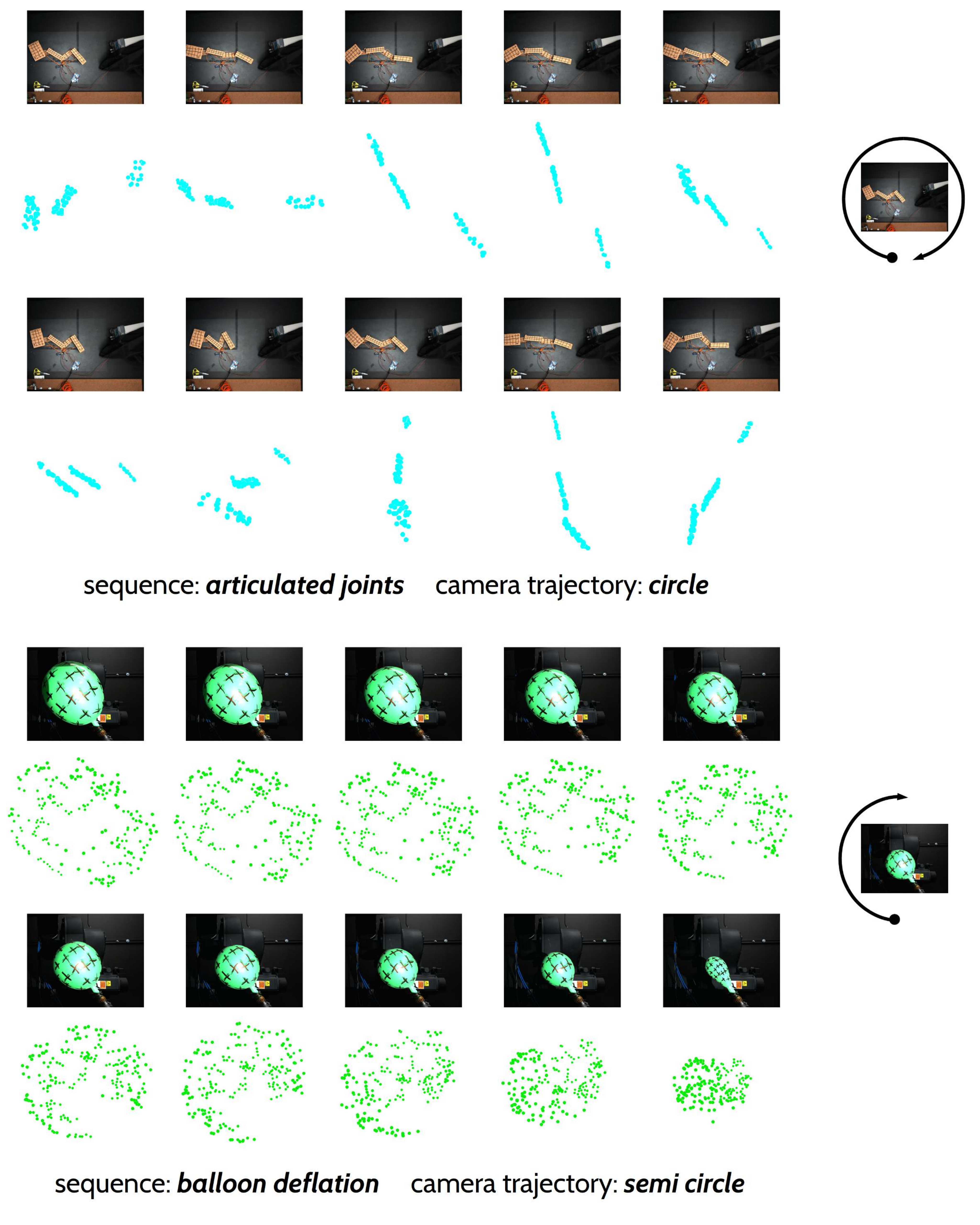} 
\caption{ Visualisation of NRSfM challenge results (\textbf{articulated joints} and \textbf{balloon deflation} sequences). We show images and the corresponding 3D reconstructions underneath. The camera trajectory is schematically visualised on the right. } 
\label{fig:challenge_2A} 
\end{figure*}

\begin{figure*}[t!] 
\centering 
  \includegraphics[width=.78\linewidth]{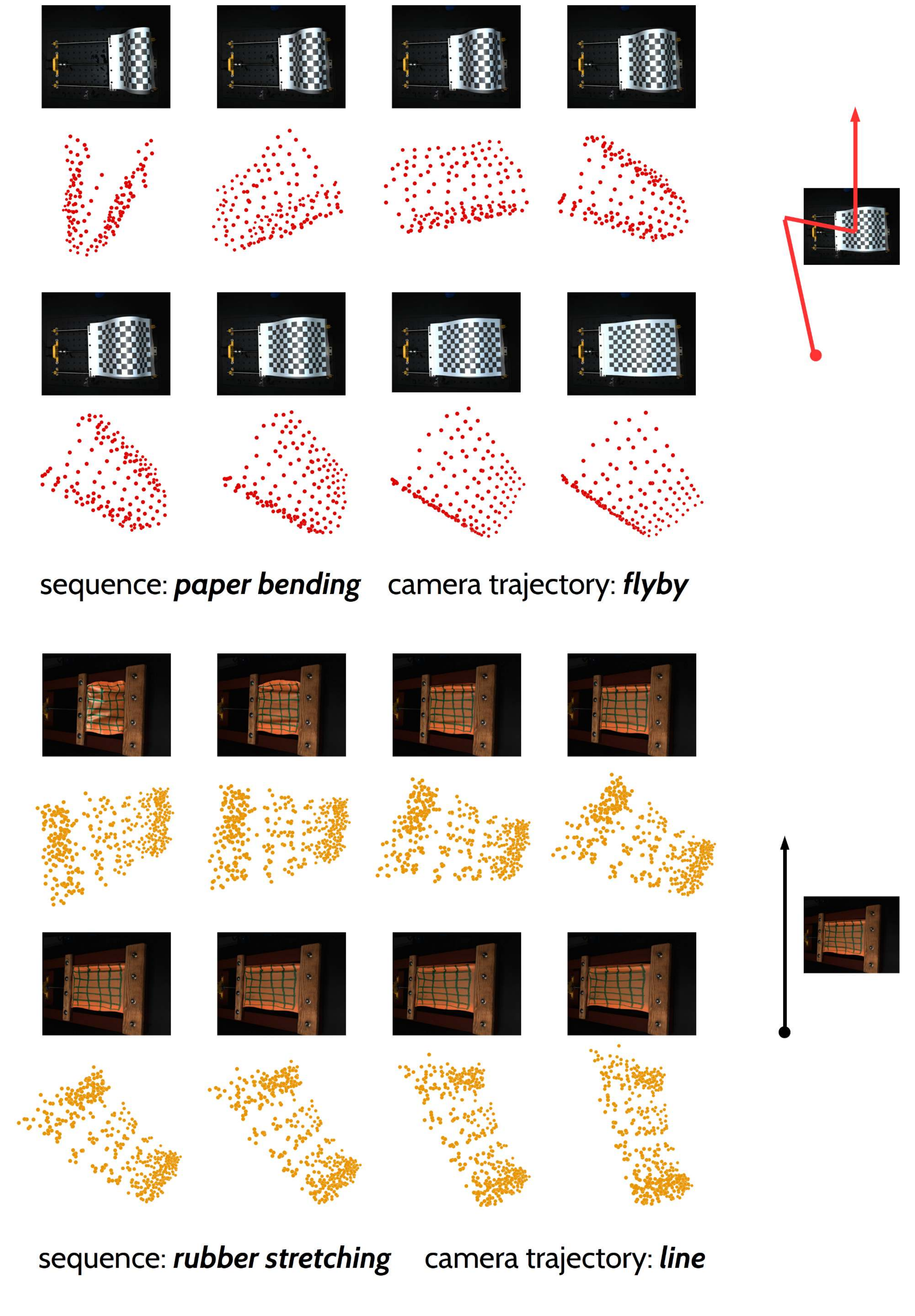} 
\caption{ Visualisation of NRSfM challenge results (\textbf{paper bending} and \textbf{rubber stretching} sequences). We show images and the corresponding 3D reconstructions underneath. The camera trajectory is schematically visualised on the right. } 
\label{fig:challenge_2B} 
\end{figure*}

\begin{figure*}%[t!] 
\centering 
  \includegraphics[width=1.0\linewidth]{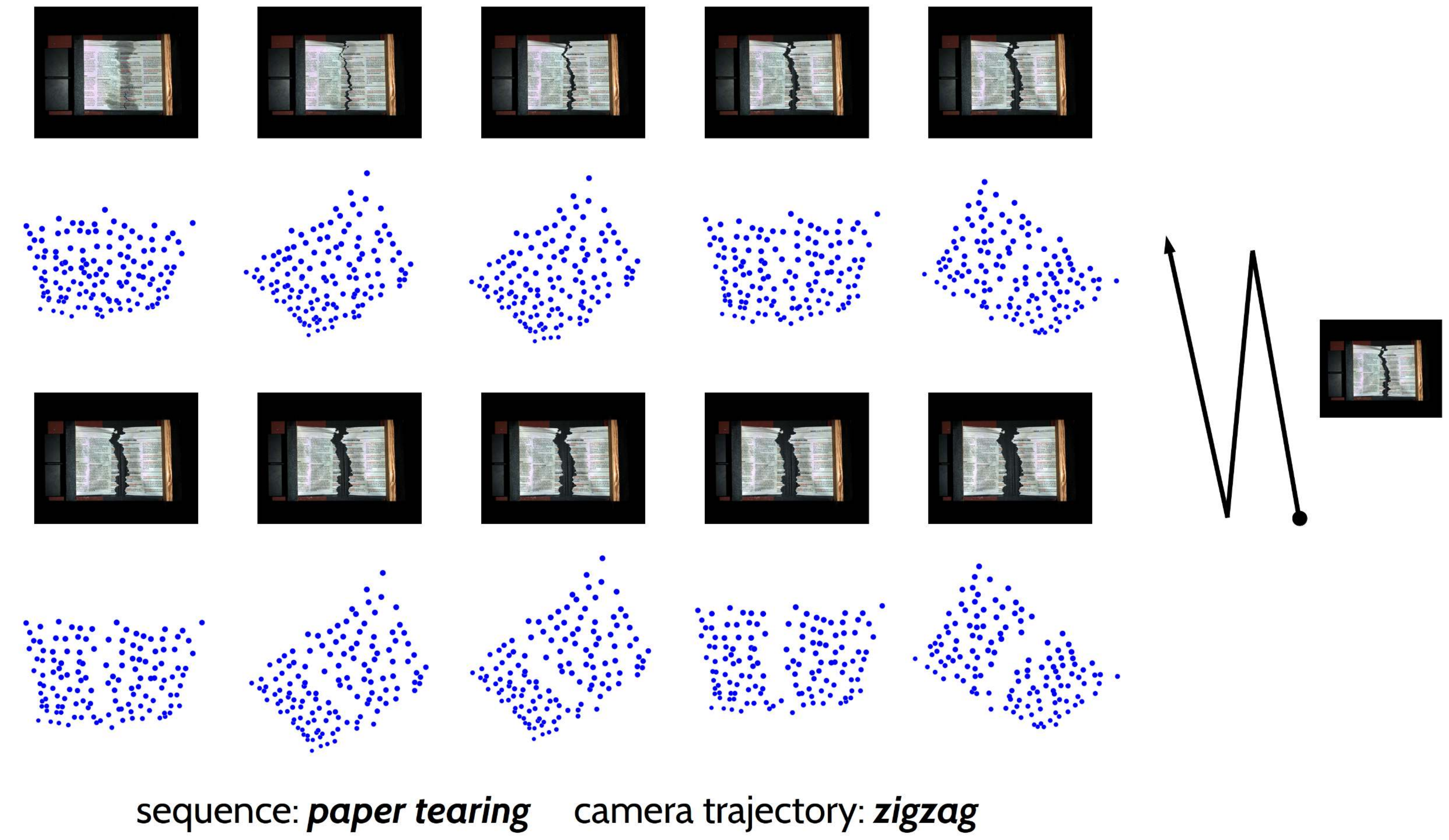} 
\caption{ Visualisation of NRSfM challenge results (\textbf{paper tearing} sequence). We show images and the corresponding 3D reconstructions underneath. The camera trajectory is schematically visualised on the right. } 
\label{fig:challenge_2C} 
\end{figure*}

The NRSfM challenge dataset contains dynamic 3D reconstructions of five different scenes with deforming objects. 
In total, there are $30$ measurement matrices obtained by applying six different virtual camera trajectories to each of the dynamic scenes. 
The sequence names consist of the scene/object and camera trajectory pairs. 
The reconstructed objects are \textbf{articulated joints} (or just \textbf{articulated}), \textbf{balloon deflation} (or just \textbf{balloon}), \textbf{paper bending} (or just \textbf{paper}), \textbf{rubber stretching} (or just \textbf{stretch}) and \textbf{paper tearing} (or just \textbf{tearing}). 
The camera trajectories are \textit{circle}, \textit{flyby}, \textit{line}, \textit{semi-circle}, \textit{tricky} and \textit{zigzag}. 
For the evaluation purposes, only the measurement matrices, as well as a single ground truth 3D state per scene, are available. 
The reconstructions are sent to the evaluation page of the NRSfM challenge and posted publicly. 
For further details, please refer to \cite{Jensen2018}. % Jensen \textit{et al.}~ 
The results obtained on the NRSfM challenge by our D-CMDR are summarised in Table \ref{tab:NRSfM_chellenge}. 
The table provides a detailed summary for every scene and camera trajectory, including the overall mean RMSE for each scene (over all camera trajectories) and each camera trajectory (over all scenes). 
D-CMDR achieves the overall mean RMSE of $50.19$ $mm$. 
The most accurate camera trajectory for D-CMDR is \textit{zigzag}, whereas the most challenging one is \textit{circle}. 
Excluding the \textit{circle}, D-CMDR improves its overall mean RMSE to $44.8$ $mm$. 
The most challenging scene for D-CMDR is \textbf{stretch}. 
All in all, D-CMDR shows scalable results, \textit{i.e.,} its average accuracy does not vary much across the scenes (the range of mean RMSE is $[37.7; 57.5]$ $mm$). 
We outperform the recent methods % over all sequences and camera trajectories 
\cite{Dai2014, Kong_2016, Chhatkuli2016, Lee2016} and come close to \cite{DelBue2010} ($48.79$ $mm$). 
% 
% For the \textit{tricky}, we obtain the m.~RMSE across all sequences of $46.74$ $mm$ which is among the best four results \cite{Jensen2018}. 
% 
The error distribution across different camera trajectories shows that a sufficient variety of angles of view around a central position is necessary for an accurate reconstruction by D-CMDR. 
Among all camera trajectories, \textit{zigzag} comes closest to the required pose variety and configuration. 
The second optimal camera trajectory is \textit{line} followed by \textit{tricky}, \textit{flyby}, \textit{semi-circle} and \textit{circle} in the descending order of mean RMSE. 
For the \textit{tricky}, we obtain the m.~RMSE across all sequences of $46.74$ $mm$ which is among the best four results reported in \cite{Jensen2018} (in total, Jensen \textit{et al.}~\cite{Jensen2018} report on $16$ NRSfM methods). 
D-CMDR reconstructs \textbf{balloon} most accurately among all five scenes. 
The similar situation is also observed for the vast majority of the methods evaluated in \cite{Jensen2018}. 
Figs.~\ref{fig:challenge_2A}--\ref{fig:challenge_2C} show exemplary reconstructions for all five scenes and five different camera trajectories. 
% 
% Even with such a simple technique, we are achieve... 

$\qquad$ $\qquad$ $\qquad$ $\qquad$ $\qquad$ $\qquad$ $\qquad$ $\qquad$

{\small
\bibliographystyle{ieee}
\bibliography{egbib}
}

\end{document}